\documentclass{ieeeaccess}
\usepackage{cite}
\usepackage{amsmath,amssymb,amsfonts}
\usepackage{algorithmic}
\usepackage{graphicx}
\usepackage{textcomp}
\usepackage{caption}

\usepackage{multirow}
\usepackage{array}

\def\BibTeX{{\rm B\kern-.05em{\sc i\kern-.025em b}\kern-.08em
    T\kern-.1667em\lower.7ex\hbox{E}\kern-.125emX}}
\begin{document}
\history{Received 8 August 2023, accepted 7 September 2023, date of publication 11 September 2023}
\doi{10.1109/ACCESS.2023.3314340}

\title{Reconstructing Continuous Light Field from Single Coded Image}
\author{\uppercase{Yuya Ishikawa}\authorrefmark{1}, \uppercase{Keita Takahashi}\authorrefmark{1}\IEEEmembership{Member, IEEE},
\uppercase{Chihiro Tsutake}\authorrefmark{1}\IEEEmembership{Member, IEEE}, \uppercase{and Toshiaki Fujii}\authorrefmark{1}\IEEEmembership{Member, IEEE}} 

\address[1]{Graduate School of Engineering, Nagoya University, Nagoya 464-8603, Japan}


\markboth
{Y. Ishikawa \headeretal: Reconstructing Continuous Light Field from Single Coded Image}
{Y. Ishikawa \headeretal: Reconstructing Continuous Light Field from Single Coded Image}

\corresp{Corresponding author: Yuya Ishikawa (e-mail: ishikawa.yuya@fujii.nuee.nagoya-u.ac.jp).}

\begin{abstract}
We propose a method for reconstructing a continuous light field of a target scene from a single observed image. Our method takes the best of two worlds: joint aperture-exposure coding for compressive light-field acquisition, and a neural radiance field (NeRF) for view synthesis. Joint aperture-exposure coding implemented in a camera enables effective embedding of 3-D scene information into an observed image, but in previous works, it was used only for reconstructing discretized light-field views. NeRF-based neural rendering enables high quality view synthesis of a 3-D scene from continuous viewpoints, but when only a single image is given as the input, it struggles to achieve satisfactory quality. Our method integrates these two techniques into an efficient and end-to-end trainable pipeline. Trained on a wide variety of scenes, our method can reconstruct continuous light fields accurately and efficiently without any test time optimization. To our knowledge, this is the first work to bridge two worlds: camera design for efficiently acquiring 3-D information and neural rendering.
\end{abstract}


\titlepgskip=-21pt

\maketitle

\section{Introduction}
\label{sec:introduction}

\PARstart{A} light field is usually represented as a set of dense multi-view images. Thanks to the abundant information contained, light fields have been used for various applications including depth estimation~\cite{honauer2017dataset,shin18epinet}, object/material recognition~\cite{Maeno2013, wang20164d}, view synthesis~\cite{Kalantari2016,mildenhall2019llff, broxton2020immersive}, and 3-D display~\cite{shin18epinetwetzstein2012tensor,Huang:2015:LightFieldStereoscope,lee2016additive}. 

Acquisition of a light field is a long-standing issue~\cite{wilburn2005high,ng2006digital,fujii2006multipoint,Taguchi2009}. Since light-field views are highly redundant with each other, view by view sampling seems to be a waste of resources. To achieve more efficient acquisition, camera-side coding schemes have been developed~\cite{veeraraghavan2007dappled,liang2008programmable,nagahara2010programmable,marwah2013compressive,Inagaki_2018_ECCV,Sakai_2020_ECCV,Guo2022TPAMI}. Combined with learning-based reconstruction algorithms, the latest methods with joint aperture-exposure coding \cite{Tateishi_2021_ICIP,vargas_2021_ICCV,Mizuno_2022_CVPR,Tateishi_2022_IEICE} drastically reduce the number of images required for high-quality reconstruction; even a single coded image alone is sufficient to reconstruct a light field (with, e.g., $5 \times 5$ views) with convincing quality. However, what is obtained from these methods is only a set of discretized views. Since there is no physical ``viewpoint grid'' in the target scene, more desirable is a continuous representation of a target 3-D scene that can be observed from continuous viewpoints. 

\begin{figure}[t]
 \begin{center}
    \includegraphics[width=\columnwidth]{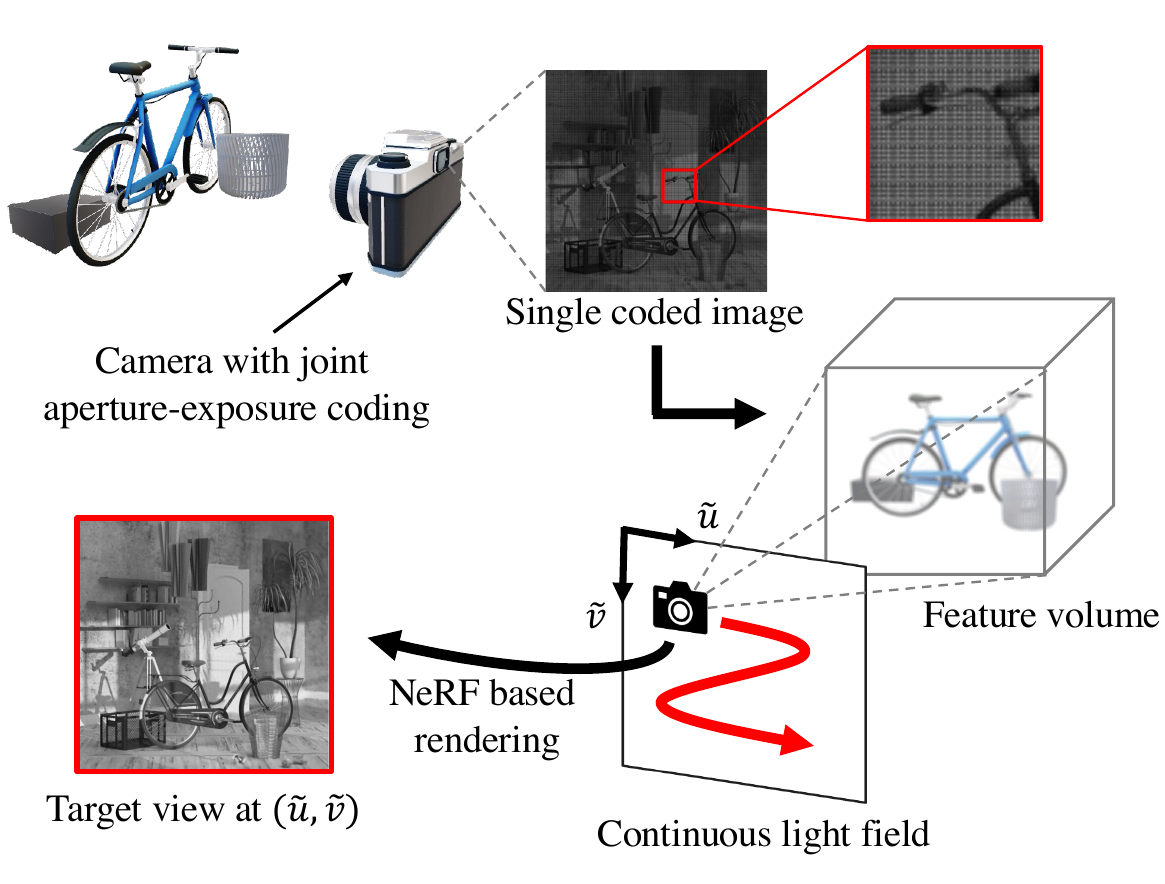}
    \caption{Our goal is to reconstruct continuous light field from single coded image alone. To this end, we integrate joint aperture-exposure coding and NeRF-based rendering.}
    \label{fig:goal}
  \end{center}
\end{figure}

Our goal is to develop a method for reconstructing a continuous light field for a target scene from a single observed image, as depicted in Fig.~\ref{fig:goal}. To this end, we take the best of two worlds: joint aperture-exposure coding~\cite{Tateishi_2021_ICIP,vargas_2021_ICCV,Mizuno_2022_CVPR,Tateishi_2022_IEICE} for compressive light-field acquisition, and a neural radiance field (NeRF) ~\cite{mildenhall2020nerf,kaizhang2020nerf++,barron2021mipnerf,Feng2021SIGNET} for view synthesis. Joint aperture-exposure coding enables effective embedding of 3-D scene information into an observed image. A NeRF-based representation enables high quality rendering of a target scene from continuous viewpoints. Our method integrates these two techniques into an efficient and end-to-end trainable pipeline. Trained on a wide variety of scenes, our method can reconstruct continuous light fields accurately and efficiently without any test time optimization. 

To the best of our knowledge, there is no previous work that can directly obtain a continuous light field from a single coded image in the context of compressive light-field acquisition. Moreover, we are the first to use not a normal (uncoded) but a coded image as the input for NeRF-based neural rendering. We show that the integration of camera-side coding and NeRF-based representation enables accurate and efficient rendering of a 3-D scene only from a single observed image. We believe our contribution will open up a new field embracing camera design and neural rendering.

The remainder of this paper is organized as follows. Section II briefly gives some background on compressive light-field acquisition and view synthesis to clarify the position of our method. Section III describes our method including the camera-side coding scheme and reconstruction of a continuous light field. Section IV presents several experimental validations including comparisons with the state-of-the-art methods. Section V concludes the paper.

\section{Background}
\label{sec:related_work}

\subsection{Compressive Light Field Acquisition}
\label{sec:comp_LF}

Traditionally, a light field was captured using an array of cameras~\cite{wilburn2005high,fujii2006multipoint,Taguchi2009} and a lenslet-based camera~\cite{adelson1992single,ng2006digital}. More recently, coded aperture cameras~\cite{veeraraghavan2007dappled,liang2008programmable,nagahara2010programmable,marwah2013compressive,Inagaki_2018_ECCV,Sakai_2020_ECCV,Guo2022TPAMI} have been developed to increase the efficiency of light field acquisition. For example, two to four images, taken with different aperture-coding patterns, are sufficient to computationally reconstruct a light field with $5\times5$ or $8\times8$ views. However, the reconstruction quality obtained with a single coded image is still unsatisfactory. Joint aperture-exposure coding~\cite{Tateishi_2021_ICIP,vargas_2021_ICCV,Mizuno_2022_CVPR,Tateishi_2022_IEICE} enables more flexible coding patterns within a single exposure time. It was demonstrated that with this advanced coding scheme, only a single image is sufficient to achieve high-quality reconstruction.

A fundamental limitation of these previous works is the discontinuity of the viewpoints; what is obtained from these methods is a set of images at discretized viewpoints. To break this limitation, our method is designed to reconstruct a continuous light field. More specifically, using a single image captured with joint aperture-exposure coding as the input, our method derives a neural representation, from which arbitrary light rays at continuous coordinates can be rendered through volume rendering.

\subsection{View Synthesis}
\label{sec:view_syn}

Given a set of posed images of a target scene, view synthesis aims to generate arbitrary views at continuous viewpoints. For this purpose, the traditional methods used 3-D meshes, voxels, point clouds, and depth maps as the 3-D representation~\cite{ Gortler1996,matusik2000image,buehler2001unstructured,Soft3DReconstruction2017ACM,Chaurasia2013DepthSA}. Recent progress of neural networks has brought more implicit 3-D representations, such as multiplane images~\cite{zhou2018stereo} and neural radiance fields (NeRF)~\cite{mildenhall2020nerf}. A NeRF is a coordinate-based representation of a target scene, from which high-quality images can be synthesized at continuous viewpoints through volume rendering. The original NeRF is scene-specific; the network model is optimized for each target scene, and it requires many images for training. The follow-up works of the original NeRF aimed at faster rendering~\cite{yu_and_fridovichkeil2021plenoxels,garbin2021fastnerf,Reiser2021ICCV}, fewer input images~\cite{mvsnerf2021ICCV,SRF2021CVPR,wang2021ibrnet,yu2021pixelnerf,Jain_2021_ICCV}, and generalization to unseen scenes~\cite{mvsnerf2021ICCV,yu2021pixelnerf,wang2021ibrnet,SRF2021CVPR, Geonerf2022CVPR}. 

Single-view view synthesis~\cite{Niklaus_TOG_2019,Wiles_2020_CVPR, single_view_mpi,VariableMPI2020ACM,Trevithick_2021_ICCV,yu2021pixelnerf,mine2021ICCV,Xu_2022_SinNeRF} refers to the extreme case of view synthesis, where only a single view is given as the input. Due to the ill-posedness of the problem, even the latest state-of-the-art methods struggle to achieve high quality rendering. 

Constructed on the framework of NeRF~\cite{mildenhall2020nerf}, our method can synthesize arbitrary views at continuous viewpoints. As a notable difference from the previous works, our method uses only a single image captured with joint aperture-exposure coding applied on the camera. An image coded in this manner can contain richer 3-D information of a target scene than a normal (uncoded) image. Moreover, our method is designed to be light-weight and generalized over a wide variety of scenes.

\section{Proposed Method}
\label{sec:approach}

\begin{figure*}[t]
 \begin{center}
    \includegraphics[width=2.0\columnwidth]{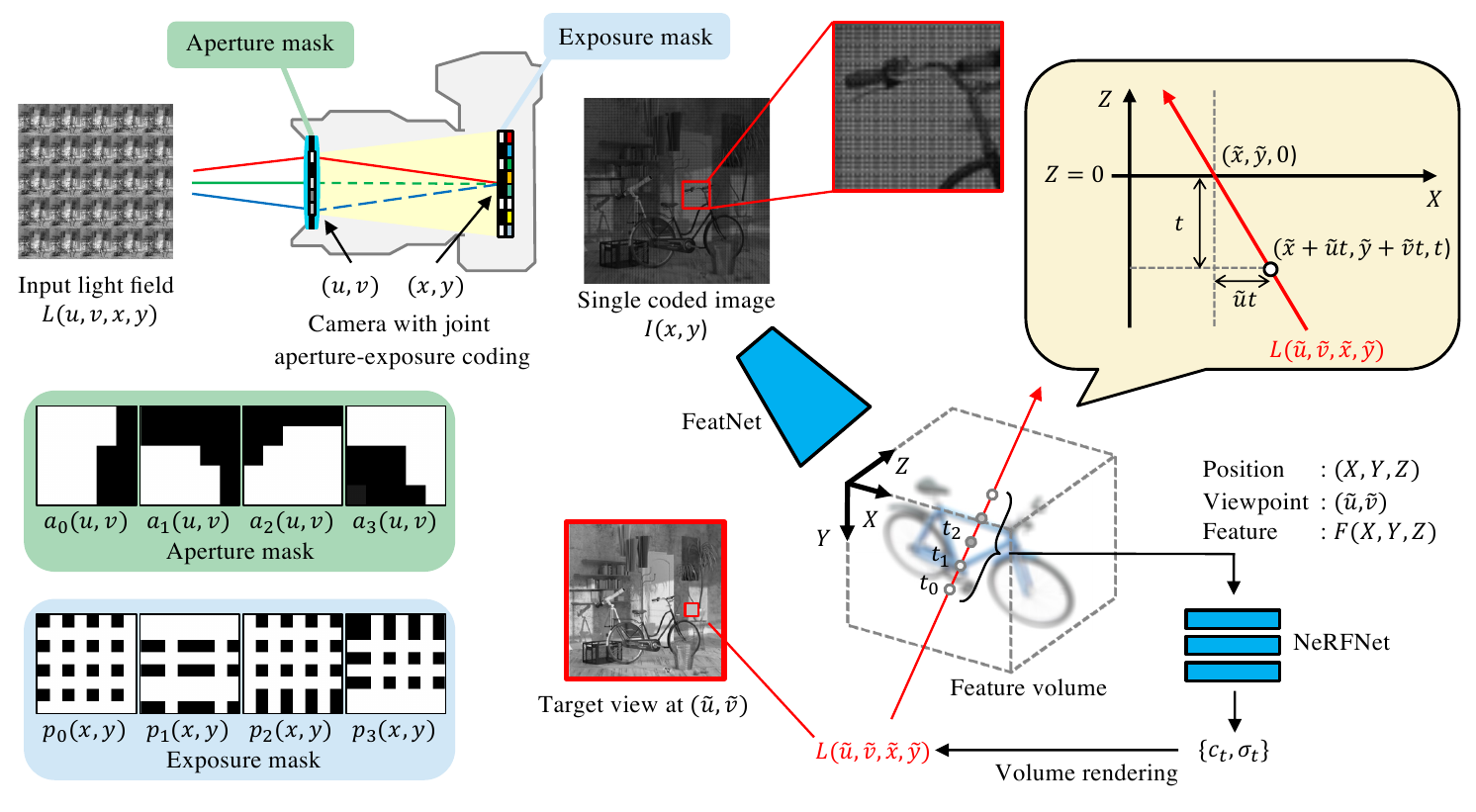}

    \caption{Overview of our method. Joint aperture-exposure coding is utilized to obtain single image used as input. FeatNet is responsible for extracting 3-D feature volume from single coded image. Using feature volume with NeRFNet, we perform volume rendering to synthesize views at arbitrary continuous viewpoints.}

    \label{fig:proposed_method}
  \end{center}
\end{figure*}

\subsection{Problem Formulation and Overview}

A light field is represented as a set of multi-view images taken from dense 2-D grid-points. It is written as $L(u,v,x,y)$, where $(u,v)$ and $(x,y)$ denote the viewpoint index and pixel positions, respectively. We also introduce a continuous light field, denoted as $L(\tilde{u},\tilde{v},\tilde{x},\tilde{y})$, where the coordinate $(\tilde{u},\tilde{v},\tilde{x},\tilde{y})$ can take arbitrary continuous values. In particular, we focus on the continuity of the viewpoints $(\tilde{u},\tilde{v})$ because this is the main focus of view synthesis. We assume that the light field is grayscale, considering the availability of imaging hardware as mentioned in \ref{sec:prop-image}.

Our goal is to reconstruct a continuous light field of a target scene from a single image alone. To fully obtain the 3-D information of the target scene, we use an image captured with joint aperture-exposure coding~\cite{Tateishi_2021_ICIP,vargas_2021_ICCV,Mizuno_2022_CVPR,Tateishi_2022_IEICE} as the input. From the coded image, we extract a feature volume that spans the target 3-D volume using a deep convolutional neural network (CNN) called FeatNet. We also construct a multi-layer perceptron (MLP) called NeRFNet to represent the radiance field of the same 3-D volume, and it is conditioned on the features extracted from the coded image. To render a view from a desired viewpoint, we make queries towards NeRFNet for the luminance and density values along the light rays that constitute the target view. Since the queries can be made at arbitrary continuous coordinates, we can reconstruct a continuous light field. Moreover,  FeatNet and NeRFNet are jointly trained and can be generalized over a wide variety of scenes. Our method can reconstruct new scenes (which are unseen during training time) accurately and efficiently without any test time optimization. An overview of our method is illustrated in Fig.~\ref{fig:proposed_method}.

\subsection{Joint Aperture-Exposure Coding}
\label{sec:prop-image}

We describe the image acquisition method for capturing a coded image that is used as the input to our method. As shown in Fig.~\ref{fig:proposed_method}, all the light rays coming into a camera are considered to constitute a light field, where the intersections of each light ray with the aperture plane and sensor plane are denoted as $(\tilde{u},\tilde{v})$ and $(\tilde{x},\tilde{y})$, respectively. Similar to the previous works, the imaging model is constructed in a discretized domain, $(u,v,x,y)$.

If the camera has no coding mechanism for incoming light rays, all the light rays reaching a single pixel $(x,y)$ are summed together to produce a pixel value. The observed image, $I(x,y)$, is given as
\begin{eqnarray}
I(x,y) = \sum_{u,v} L(u,v,x,y).
\label{eq:ordinary}
\end{eqnarray}
Note that variations along $(u,v)$ dimensions are simply blurred out in $I(x,y)$, making them difficult to recover.

Coded aperture cameras~\cite{veeraraghavan2007dappled,liang2008programmable,nagahara2010programmable,marwah2013compressive,Inagaki_2018_ECCV,Guo2022TPAMI} have been used to effectively encode $(u,v)$ dimensions. Specifically, a semi-transparent mask, $a(u,v)$, is inserted at the aperture plane to encode the incoming light rays. Each pixel value  is the weighted sum of light rays, described as
 \begin{eqnarray}
I(x,y) = \sum_{u,v} a(u,v) L(u,v,x,y).
\label{eq:ca}
\end{eqnarray}
However, it is difficult to accurately reconstruct  the light field from a single coded image. Therefore, two or more images taken with different mask patterns are used for better reconstruction quality.

To embed more abundant information into a single observed image, joint aperture-exposure coding~\cite{Tateishi_2021_ICIP,vargas_2021_ICCV,Mizuno_2022_CVPR,Tateishi_2022_IEICE} has been investigated recently. Within a single exposure time, both the aperture mask and pixel-wise exposure mask are synchronously controlled to encode the light rays. The image formation model is written as
 \begin{eqnarray}
I(x,y) = \sum_{u,v} \left\{ \sum_{k<K} {a_k(u,v)p_k(x,y)} \right\} L(u,v,x,y)
\label{eq:ca+ce}
\end{eqnarray}
where $a_k(u,v)$ and $p_k(x,y)$ are the $k$-th aperture and exposure coding patterns, respectively, and $K$ is the number of coding patterns along time. Thanks to the complex coding scheme, a single coded image alone is sufficient for accurate reconstruction of the light field. We adopt Eq.~(\ref{eq:ca+ce}) as the image acquisition model to take the input coded image for our method.

\begin{figure*}[t]
 \begin{center}
    \includegraphics[width=1.95\columnwidth]{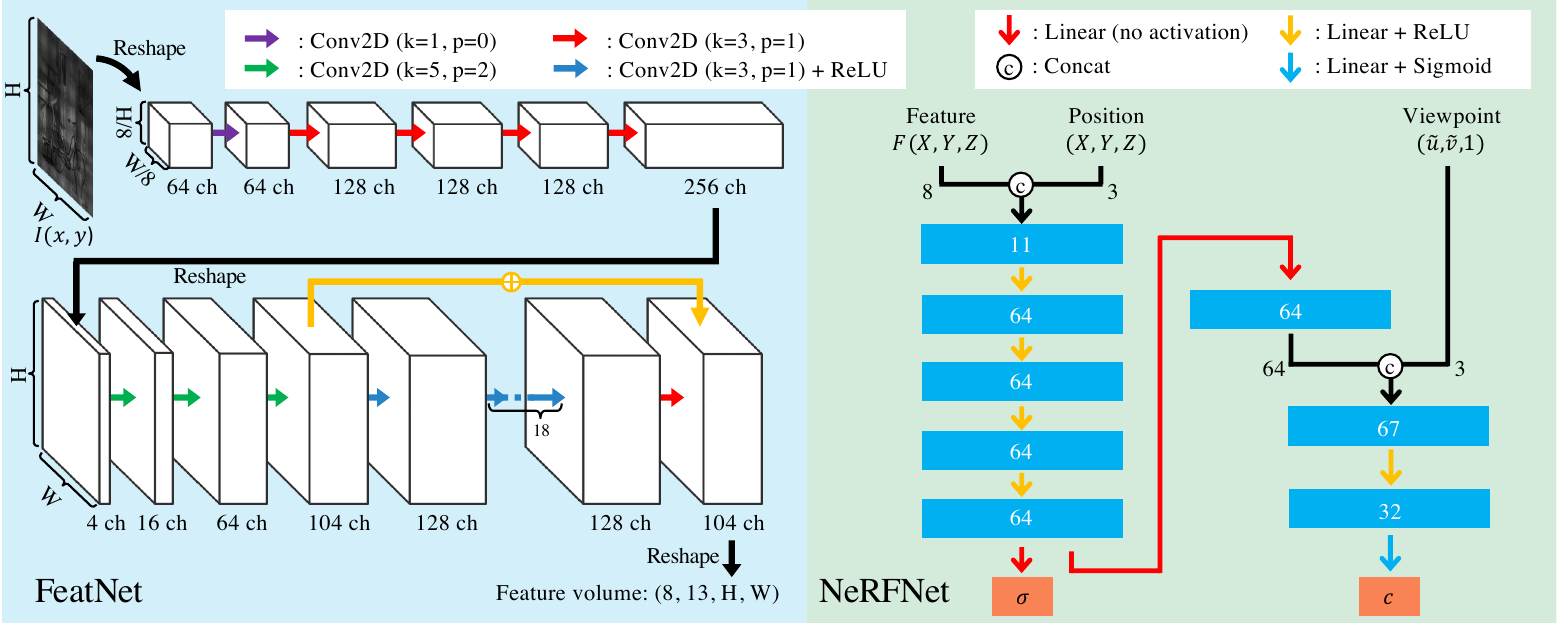}

    \caption{Network architectures for FeatNet (left) and NeRFNet (right); k and p mean kernel size and padding.}

    \label{fig:network}
  \end{center}
\end{figure*}

Joint aperture-exposure coding is not easy to implement in real camera hardware. As far as we know, only Mizuno et al.~\cite{Mizuno_2022_CVPR} demonstrated a hardware implementation that can achieve Eq.~(\ref{eq:ca+ce}), but there were some additional restrictions; the image sensor was grayscale, and $p_k(x,y)$ can take only limited patterns. Although their method was originally designed for time-varying scenes, it is also effective for static scenes. We strictly follow the design of Mizuno et al.'s working prototype. We consider grayscale light fields with $5 \times 5$ views, set $K$ to 4, and adopt the same coding patterns for $a_k(x,y)$ and $p_k(x,y)$ as Mizuno et al.'s. The mask patterns we used are shown in Fig.~\ref{fig:proposed_method}. Accordingly, our method is based on a real existing camera, rather than an idealized hypothetical camera model.

Our goal is to obtain a continuous light field of the target scene, $L(\tilde{u},\tilde{v},\tilde{x},\tilde{y})$, from a single coded image, $I(x,y)$. This is different from Mizuno et al.~\cite{Mizuno_2022_CVPR} in which the light field was reconstructed only in the discretized domain as $L(u,v,x,y)$. To achieve our goal, we integrate the idea of neural radiance fields~\cite{mildenhall2020nerf}, which enables rendering from continuous viewpoints, into the framework of compressive light-field acquisition, as detailed in the next subsection.

\subsection{Reconstruction of Continuous Light Field}

As shown in Fig.~\ref{fig:proposed_method}, our method achieves continuous light field reconstruction using two networks: FeatNet and NeRFNet; refer to Fig.~\ref{fig:network} for the detailed architectures.

We first mention a network for feature extraction, called FeatNet. This is a deep CNN that takes a single coded image $I \in {\cal R}^{H \times W}$ as the input and generates a feature volume $F\in {\cal R}^{H \times W \times D \times C}$ as the output.
\begin{eqnarray}
F = \mbox{FeatNet}(I)
\end{eqnarray}
The network architecture for FeatNet is almost the same as the one that Mizuno et al.~\cite{Mizuno_2022_CVPR} used for discretized light-field reconstruction. In our method, $F$ is interpreted as a 3-D volume with $D$ depth levels that spans the target scene, and each voxel takes a $C$-channel feature vector.  We set $D=13$ and $C=8$. The 3-D volume is re-parameterized at the normalized device coordinate (NDC) with $X,Y,Z \in [-1, 1]$ and treated as being continuous. We use trilinear interpolation to enable the querying for the feature vector $\mathbf{f}$ at a continuous 3-D coordinate $(X,Y,Z)$. We simply describe the query operation as $\mathbf{f} = F(X,Y,Z)$. 
 
We define a neural radiance field (NeRF) for the same $(X,Y,Z)$ volume. It is implemented as a multi-layer perceptron (MLP) that takes the 3-D position $(X,Y,Z)$, angle $(\theta, \phi)$, and feature vector $\mathbf{f}$ as the input and produces the luminance ($c$) and density ($\sigma$) as the output.
\begin{eqnarray}
c, \sigma &=& \mbox{NeRFNet}(X,Y,Z,\theta,\phi, \mathbf{f})
\end{eqnarray}
Different from the original NeRF~\cite{mildenhall2020nerf}, our NeRFNet takes as input the feature vector $\mathbf{f}$ extracted from the observed image. Since scene-specific information is given as the feature vector, the network weights of NeRFNet no longer need to be scene-specific but can be generalized over a wide variety of scenes. Moreover, we use a more light-weight MLP than the original NeRF. We found that, provided the feature vector as the input, a small MLP is sufficient to achieve high quality rendering.

Using the features from scene observation is not a new idea in itself; similar ideas have been used for both generalized and scene-specific NeRF-like representations~\cite{wang2021ibrnet,mvsnerf2021ICCV,yu2021pixelnerf,Kosiorek2021NeRFVAEAG,Chan_2021_CVPR,Geonerf2022CVPR}. However, our method is different from these works in that the features are extracted from not multiple images but a single image alone, and the image is acquired with a camera-side coding process to fully capture the 3-D scene information.

We finally mention how we can obtain a light field from the feature volume $F$ and NeRFNet. As shown in Fig.~\ref{fig:proposed_method}, the continuous 3-D space $(X,Y,Z)$ is associated with the continuous light field coordinate $(\tilde{u}, \tilde{v}, \tilde{x}, \tilde{y})$ via a plane+angle representation; $(\tilde{x}, \tilde{y})$ denotes the position on $Z=0$, and $(\tilde{u},\tilde{v})$ is considered the angle $(\theta,\phi)$. A light ray parameterized with $(\tilde{u}, \tilde{v}, \tilde{x}, \tilde{y})$ is mapped to $(X,Y,Z)$ as
\begin{eqnarray}
\left[
\begin{array}{c}
X \\ Y \\ Z
\end{array}
\right]
= {\cal M}_t(\tilde{u}, \tilde{v}, \tilde{x}, \tilde{y}) = 
\left[
\begin{array}{c}
\mbox{norm}(\tilde{x}+\tilde{u} t) \\ \mbox{norm}(\tilde{y}+\tilde{v}t) \\ \mbox{norm}(t)
\end{array}
\right]
\label{eq:mapping}
\end{eqnarray}
where $\mbox{norm}()$ denotes the normalization to the NDC, and $t \in [t_{\min}, t_{\max}]$ denotes a sampling position on the light ray. Note that $t_{\min}$ and $t_{\max}$ correspond to the minimum and maximum disparities allowable for the target scene.  We set $t_{\min} = -3$ and $t_{\max} = 3$. 

Using the relation of Eq.~(\ref{eq:mapping}), we can query $\mathbf{f}_t$, $c_t$, and $\sigma_t$ for a sampling position $t$ along a light ray parameterized with $(\tilde{u}, \tilde{v}, \tilde{x}, \tilde{y})$.
\begin{eqnarray}
\mathbf{f}_t &=& F({\cal M}_t(\tilde{u}, \tilde{v}, \tilde{x}, \tilde{y})), \\
c_t, \sigma_t &=& \mbox{NeRFNet}({\cal M}_t(\tilde{u}, \tilde{v}, \tilde{x}, \tilde{y}), \tilde{u},\tilde{v},\mathbf{f}_t).
\end{eqnarray}
For the values of $t$, we take 16 random stratified samples for the training time and 32 uniform samples for the test time. 
Given a set of samples along the ray, $\{ c_t, \sigma_t \}$, we perform volume rendering, which is also used in the original NeRF~\cite{mildenhall2020nerf}, to obtain the luminance of the light ray. 
\begin{eqnarray}
L(\tilde{u}, \tilde{v}, \tilde{x}, \tilde{y}) = \mbox{VolumeRendering}(\{ c_t, \sigma_t \})
\end{eqnarray}
Since the coordinate $(\tilde{u}, \tilde{v}, \tilde{x}, \tilde{y})$ is continuous, we can render arbitrary light rays; thus, we can reconstruct the continuous light field of a target scene.

\subsection{Training}

 We train FeatNet and NeRFNet in an end-to-end manner so that the original light field captured by the camera can be accurately reconstructed from the feature volume and NeRFNet. As the training data, we use discretized light fields with $5 \times 5$ views. Therefore, the training loss (we use MSE loss) is computed only for the original $5 \times 5$ viewpoints. However, thanks to the randomized sampling points along the rays (i.e., $t$ is drawn at random in the volume rendering process), our method is optimized over the continuous 3-D space rather than the discretized 3-D grid-points. Moreover, our method is not scene-specific (not requiring per-scene training) but trained over a wide variety of scenes to obtain a generalization capability; once the training has been completed, no test time optimization is necessary for new scenes.

\begin{table*}[t]
  \begin{center}
    \caption{Quantitative evaluation for compressive light-field acquisition on BasicLFSR test dataset. Reported scores are PSNR/SSIM; greater is better for both of them. Note that our method can reconstruct \textbf{continuous} light fields, whereas others~\cite{Inagaki_2018_ECCV,Guo2022TPAMI,Mizuno_2022_CVPR} can obtain only \textbf{discretized} $5\times 5$ views. Our method achieved comparable quality to Mizuno et al.'s~\cite{Mizuno_2022_CVPR} for these discretized viewpoints.} 
    \label{tab:main_table}
    \vspace{3mm}
    \begin{tabular*}{\linewidth}{wl{0.2\linewidth}||wc{0.105\linewidth}wc{0.105\linewidth}wc{0.105\linewidth}wc{0.105\linewidth}wc{0.105\linewidth}| wc{0.105\linewidth}}
      Method
      & EPFL & HCI (new) & HCI (old) & INRIA & Stanford & ALL \\ \hline
      \hline
      Inagaki~\cite{Inagaki_2018_ECCV} & 29.95/0.913 & 27.85/0.784 & 34.20/0.906 & 31.25/0.918 & 25.12/0.815 & 29.67/0.867 \\
      Guo~\cite{Guo2022TPAMI} & 30.69/0.920 & 28.49/0.809 & 33.14/0.891 & 33.45/0.938 & 25.78/0.847 & 30.31/0.881 \\
      Mizuno~(pre-trained)~\cite{Mizuno_2022_CVPR} & 32.86/0.932 & 30.86/0.841 & 37.38/0.941 & 35.02/0.939 & \textbf{29.27}/\textbf{0.899} & 33.08/0.910 \\
      Mizuno~(re-trained)\cite{Mizuno_2022_CVPR} & \textbf{33.50}/\textbf{0.939} & \textbf{31.08}/\textbf{0.844} & \textbf{38.02}/\textbf{0.947} & \textbf{35.53}/\textbf{0.942} & 29.18/0.891 & \textbf{33.46}/\textbf{0.913} \\ \hline
      
      Ours & 32.23/0.926 & 30.82/0.837 & 37.49/0.941 & 34.03/0.932 & 29.10/0.888 & 32.73/0.905 \\
      Ours (no-coding) & 29.02/0.892 & 27.02/0.760 & 32.69/0.879 & 30.45/0.911 & 23.16/0.773 & 28.47/0.843 \\
      Ours (center-only) & 27.14/0.860 & 25.98/0.742 & 31.49/0.856 & 28.47/0.881 & 21.65/0.718 & 26.95/0.812 \\ \hline
      
    \end{tabular*}

  \end{center}

\end{table*}

\newcommand{\size}{0.16\hsize}

\begin{figure*}[t]
{\tabcolsep = 0.8mm
{\renewcommand\arraystretch{0.5}
      \begin{tabular}{cccccc}
          \includegraphics[width = .15\linewidth]{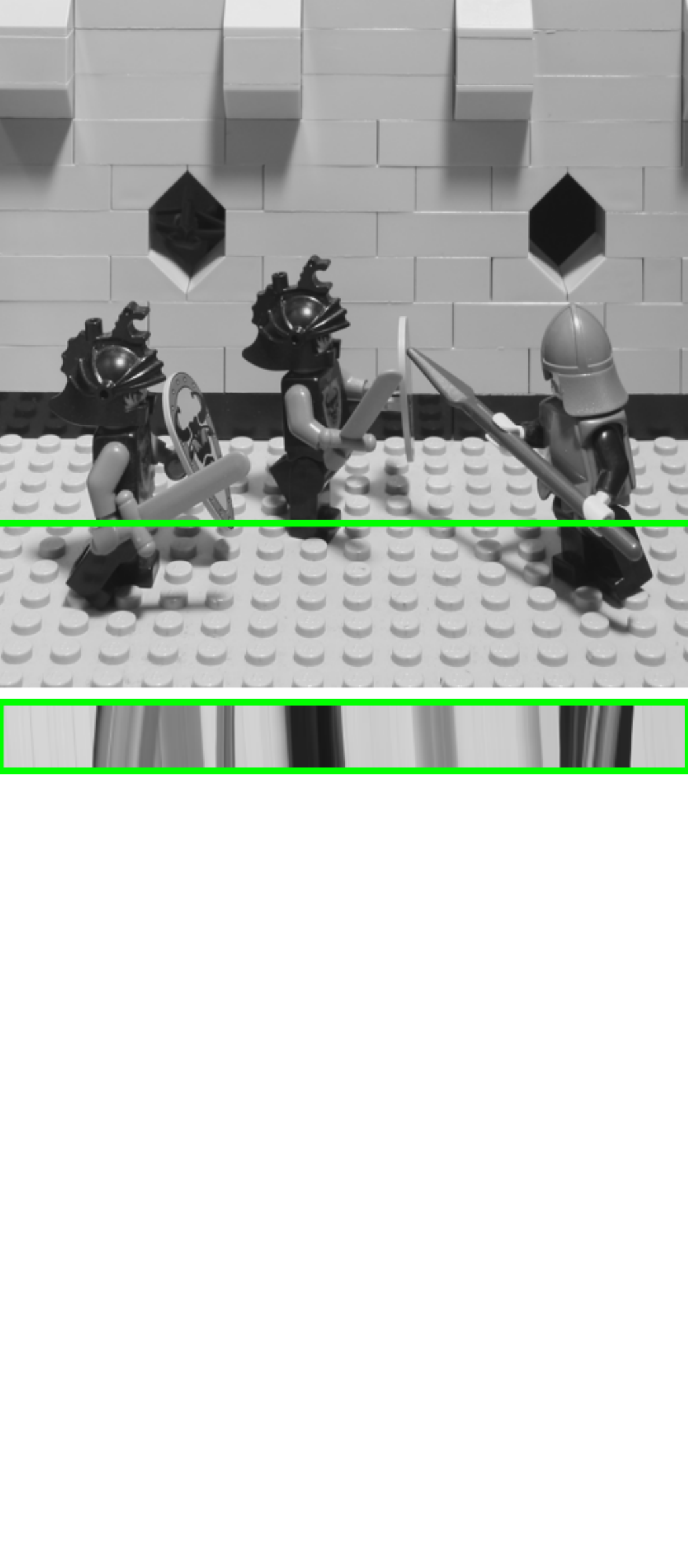} & 
          \includegraphics[width = .15\linewidth]{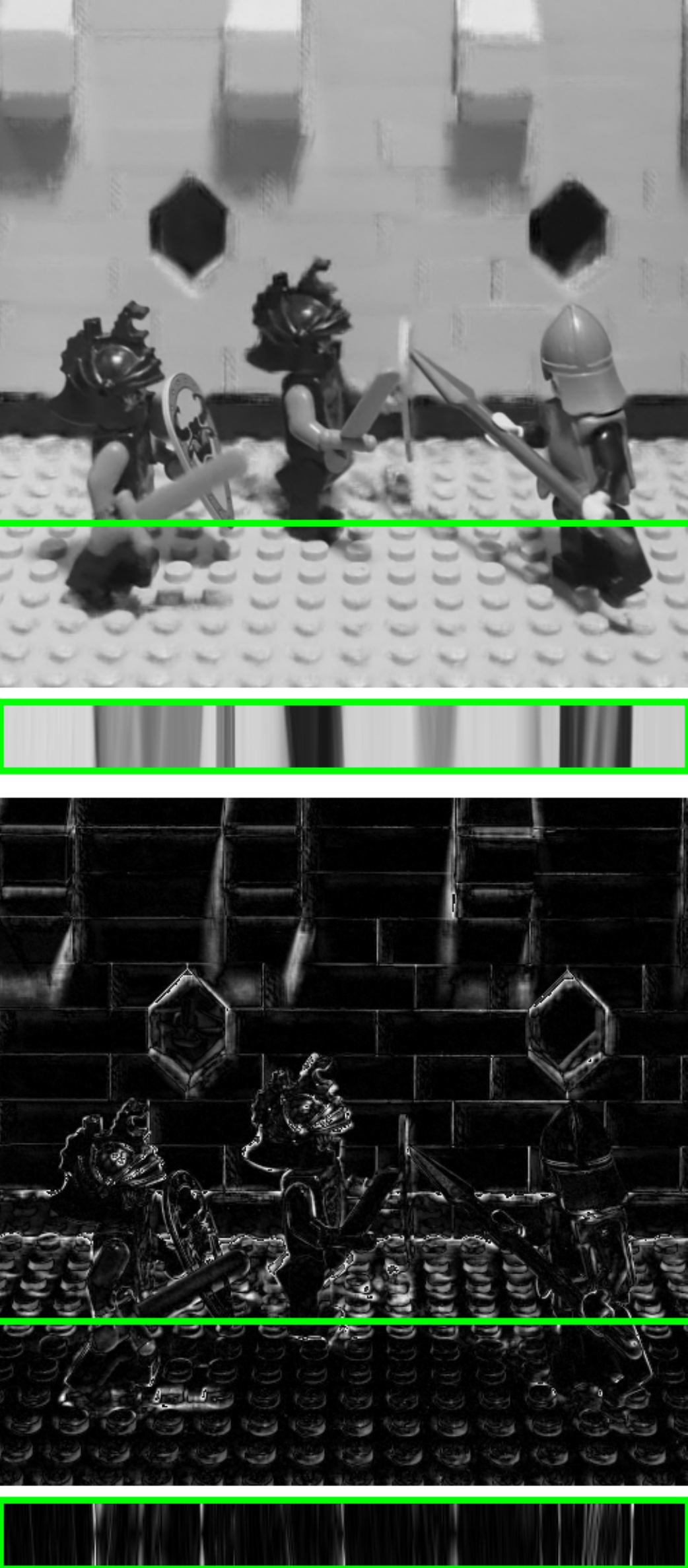} &
          \includegraphics[width = .15\linewidth]{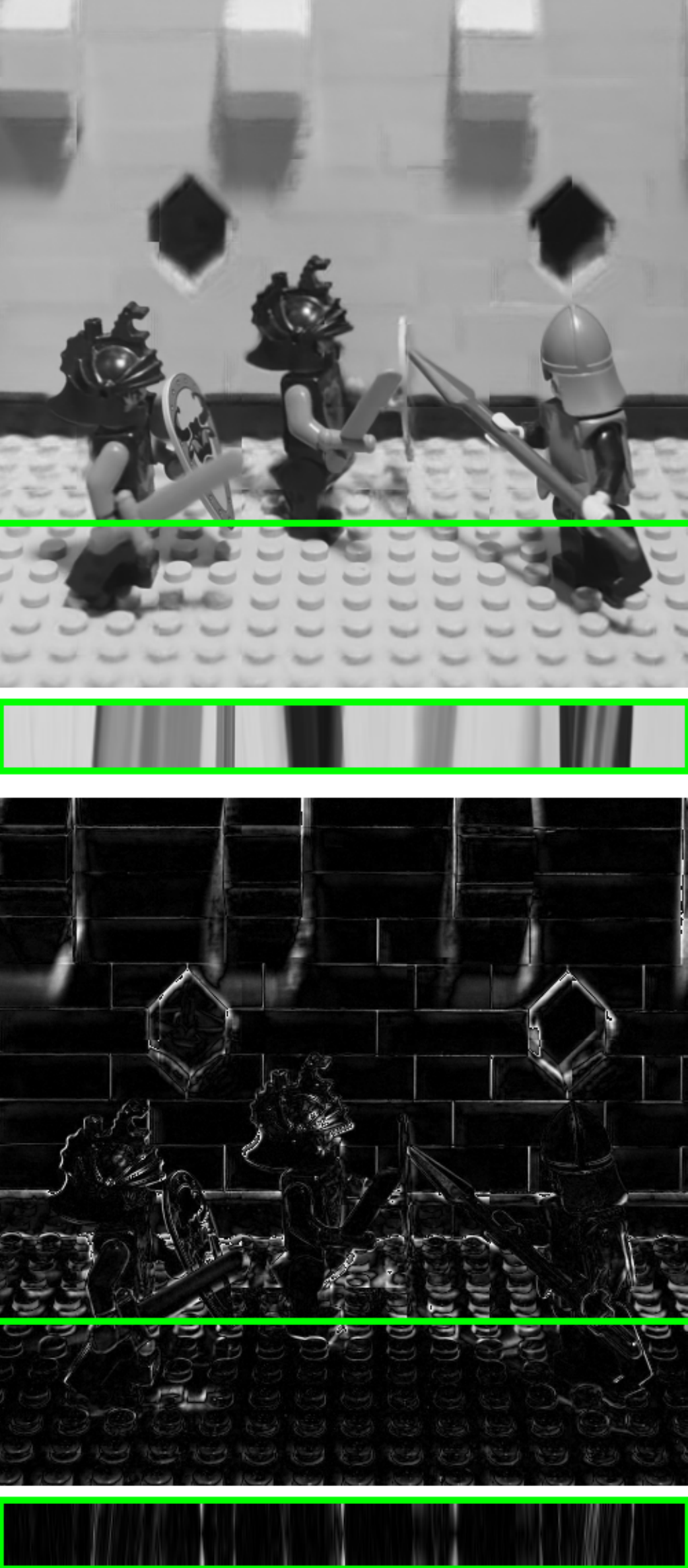} &
          \includegraphics[width = .15\linewidth]{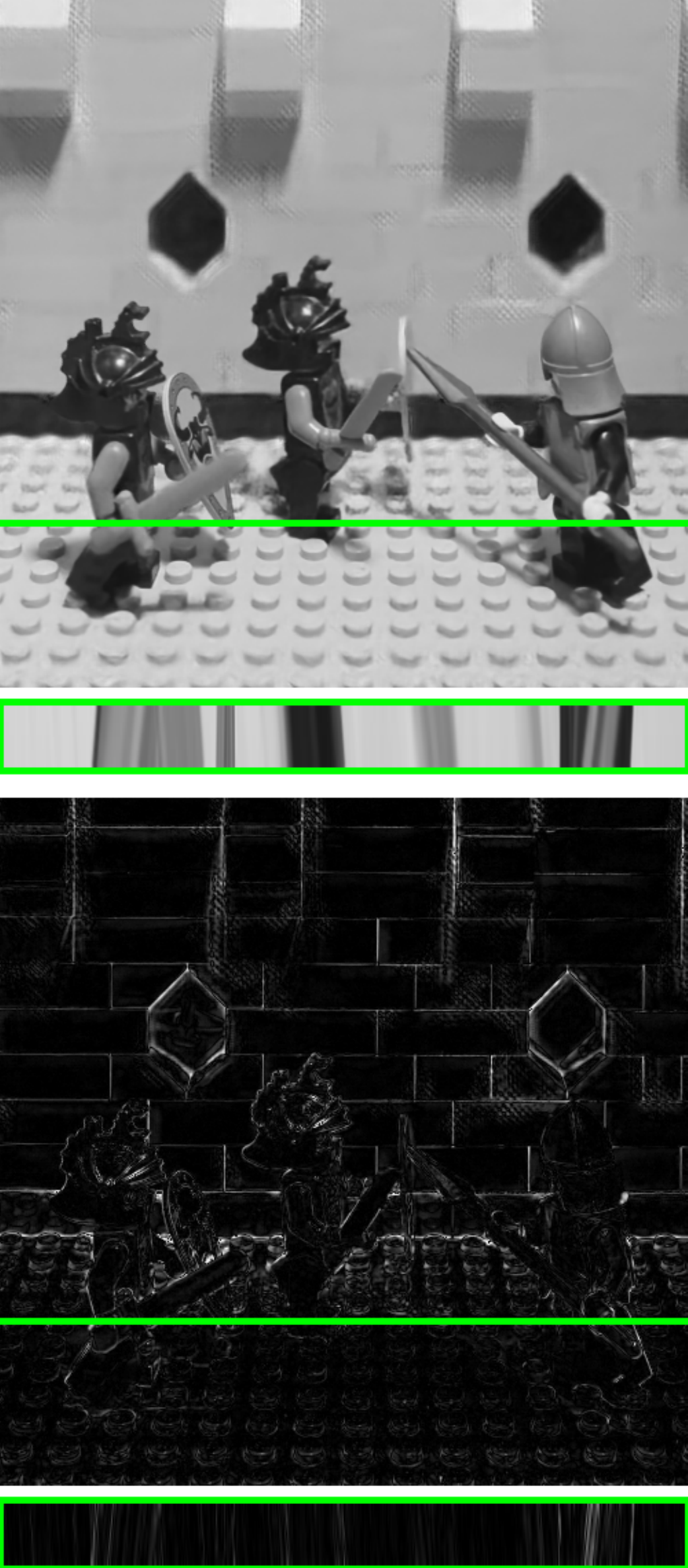} &
          \includegraphics[width = .15\linewidth]{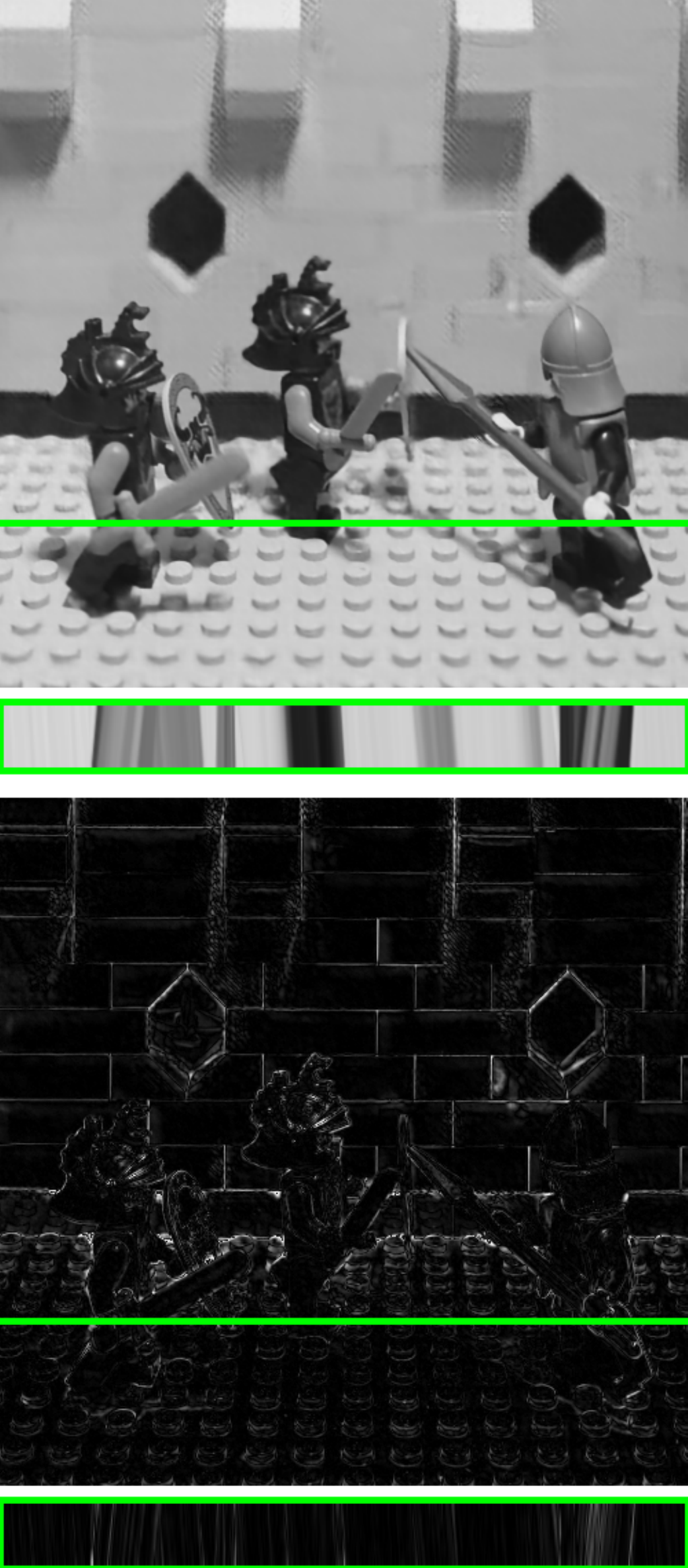} &
          \includegraphics[width = .15\linewidth]{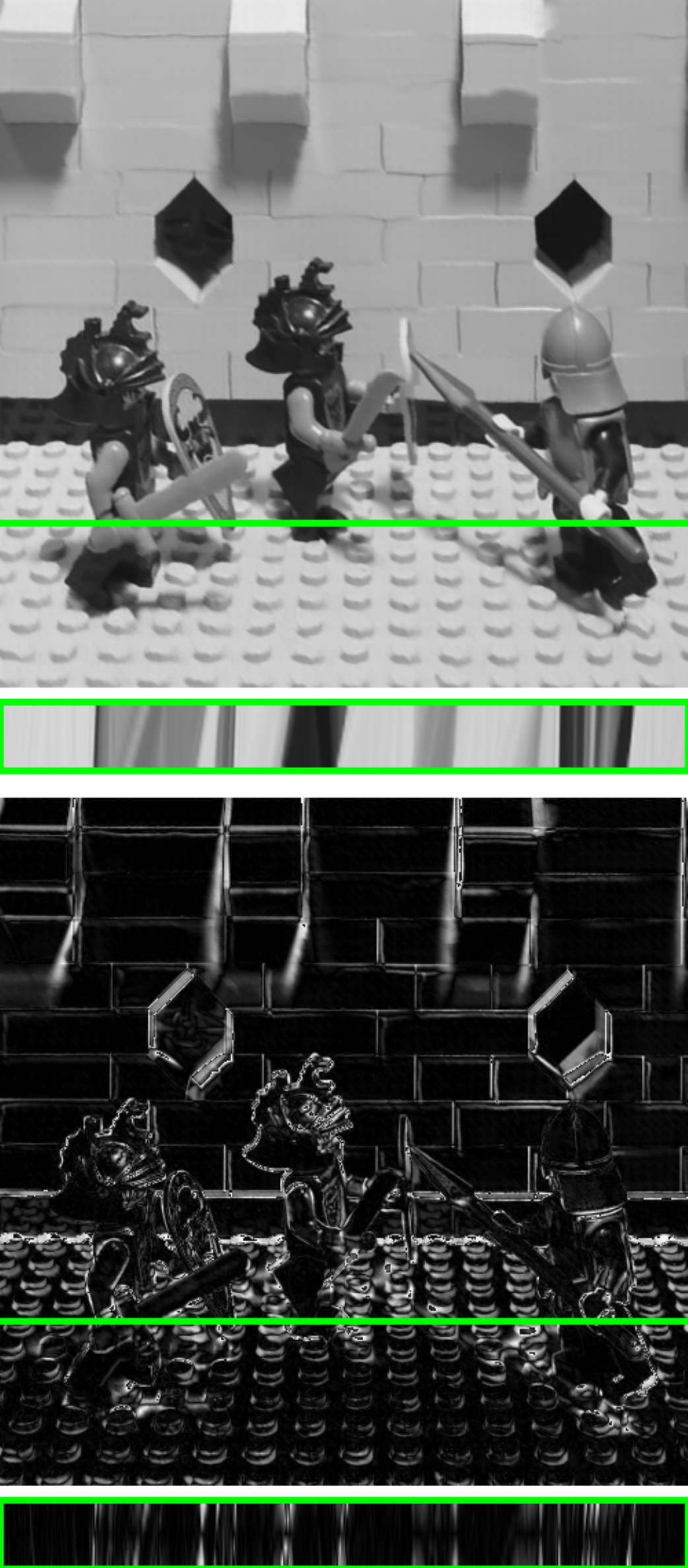} \\ \\

          \includegraphics[width = .15\linewidth]{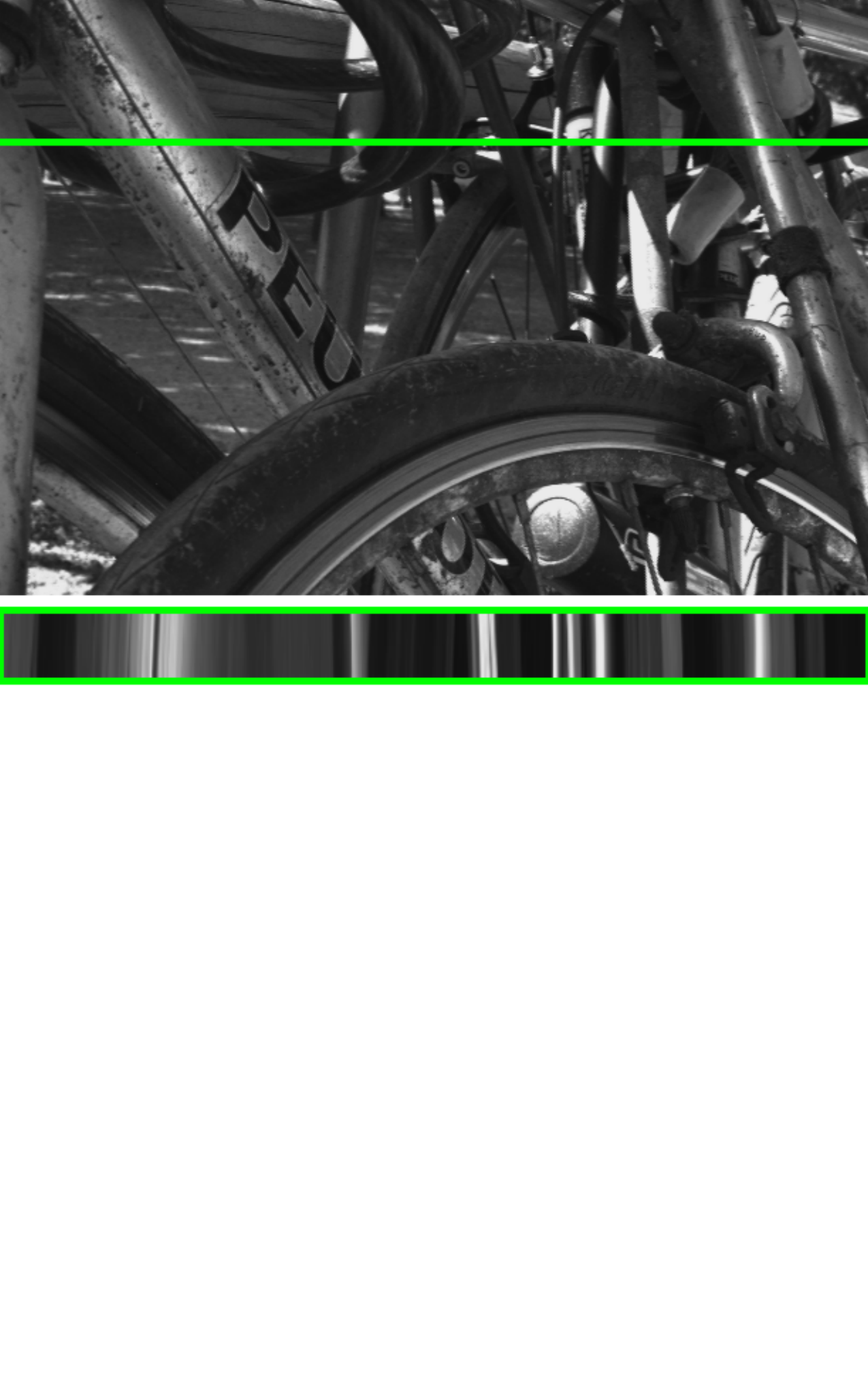} & 
          \includegraphics[width = .15\linewidth]{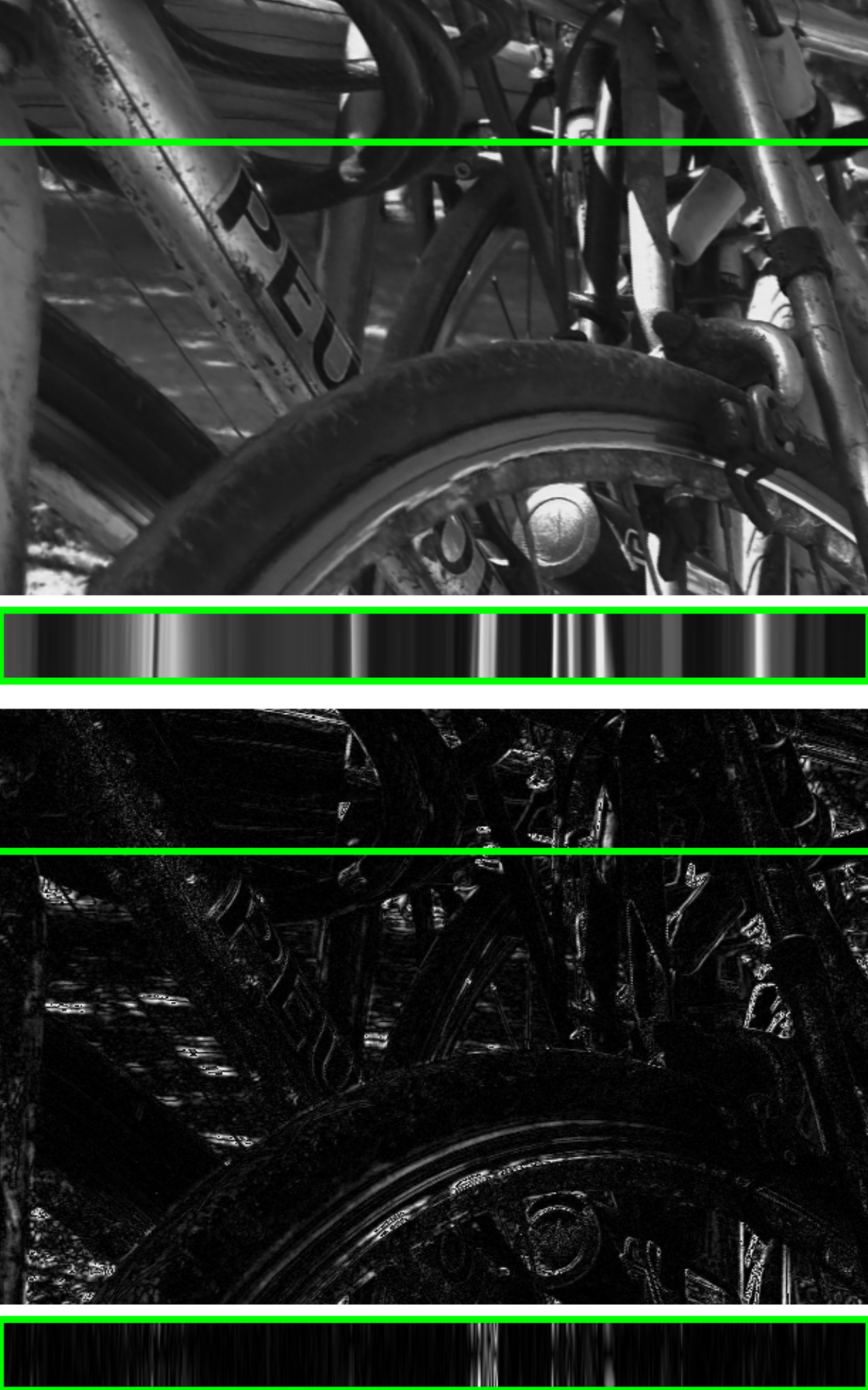} &
          \includegraphics[width = .15\linewidth]{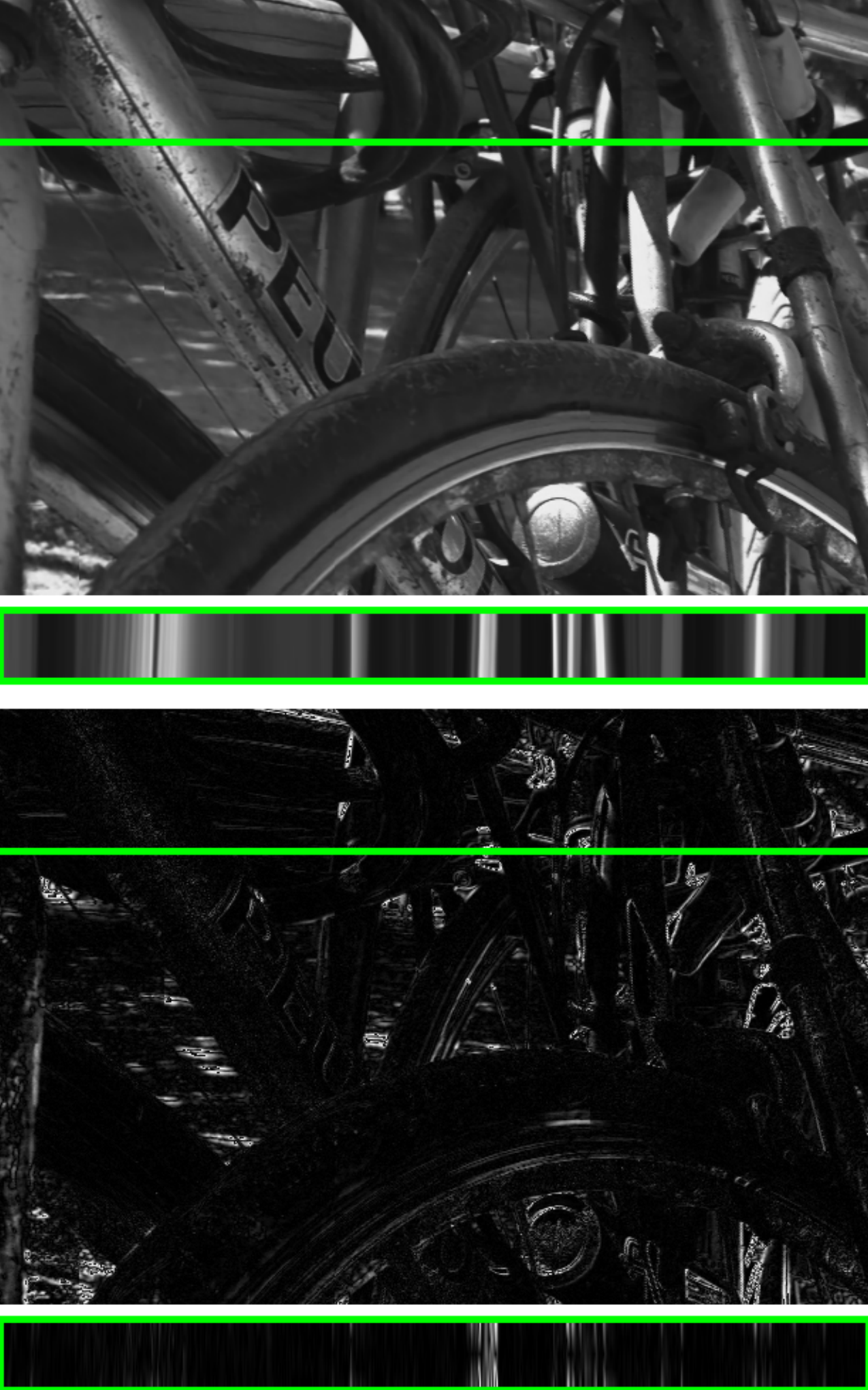} &
          \includegraphics[width = .15\linewidth]{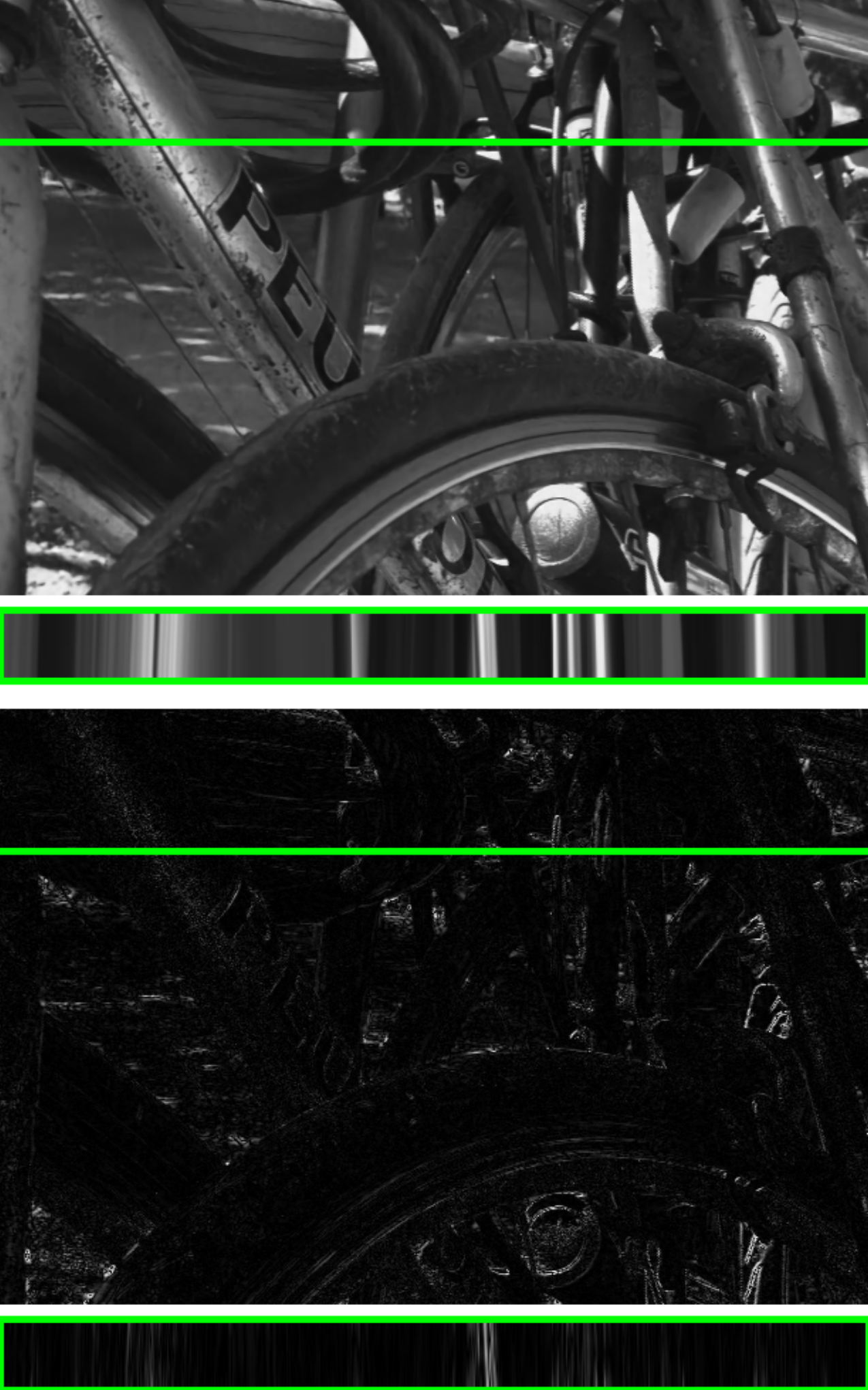} &
          \includegraphics[width = .15\linewidth]{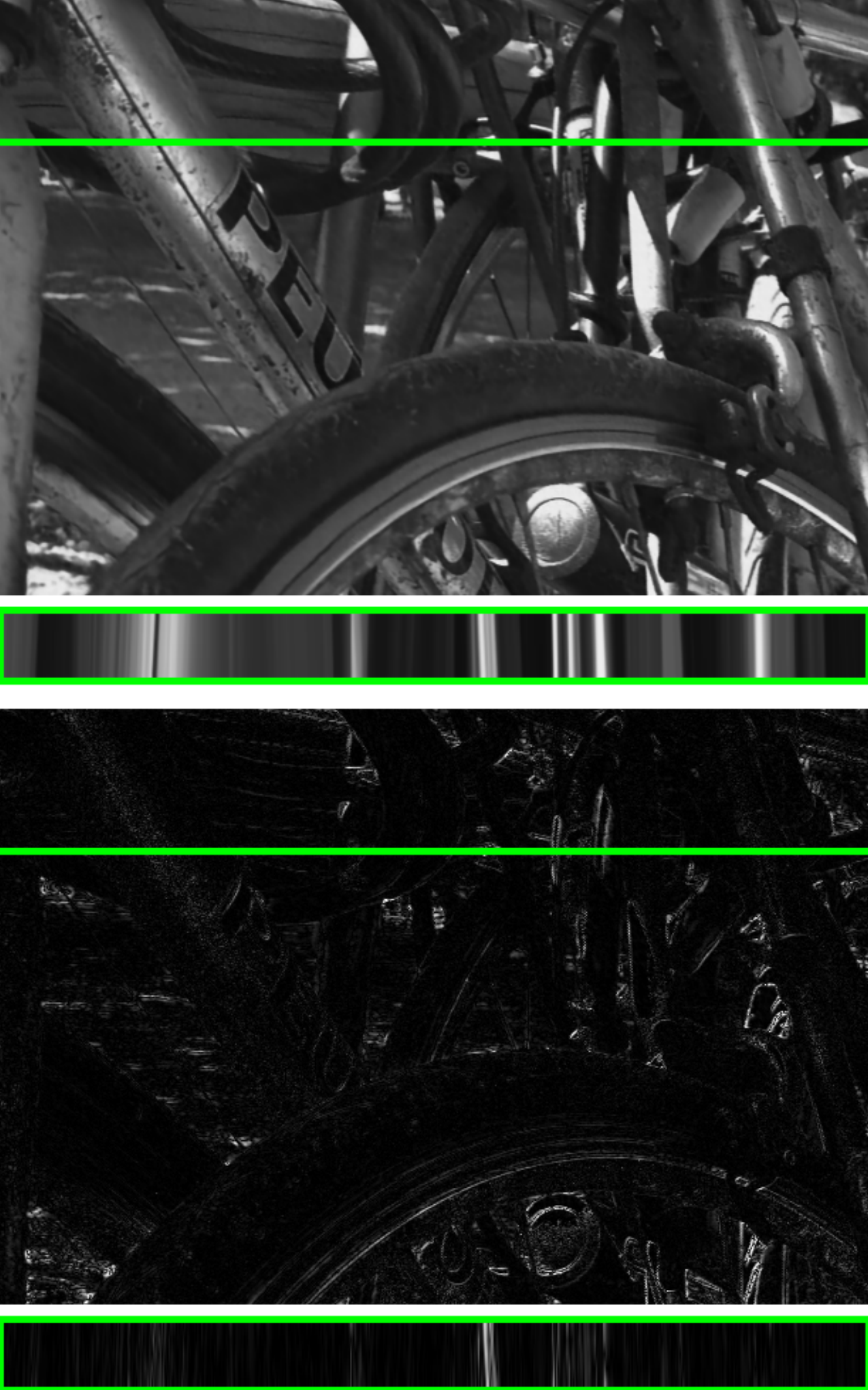} &
          \includegraphics[width = .15\linewidth]{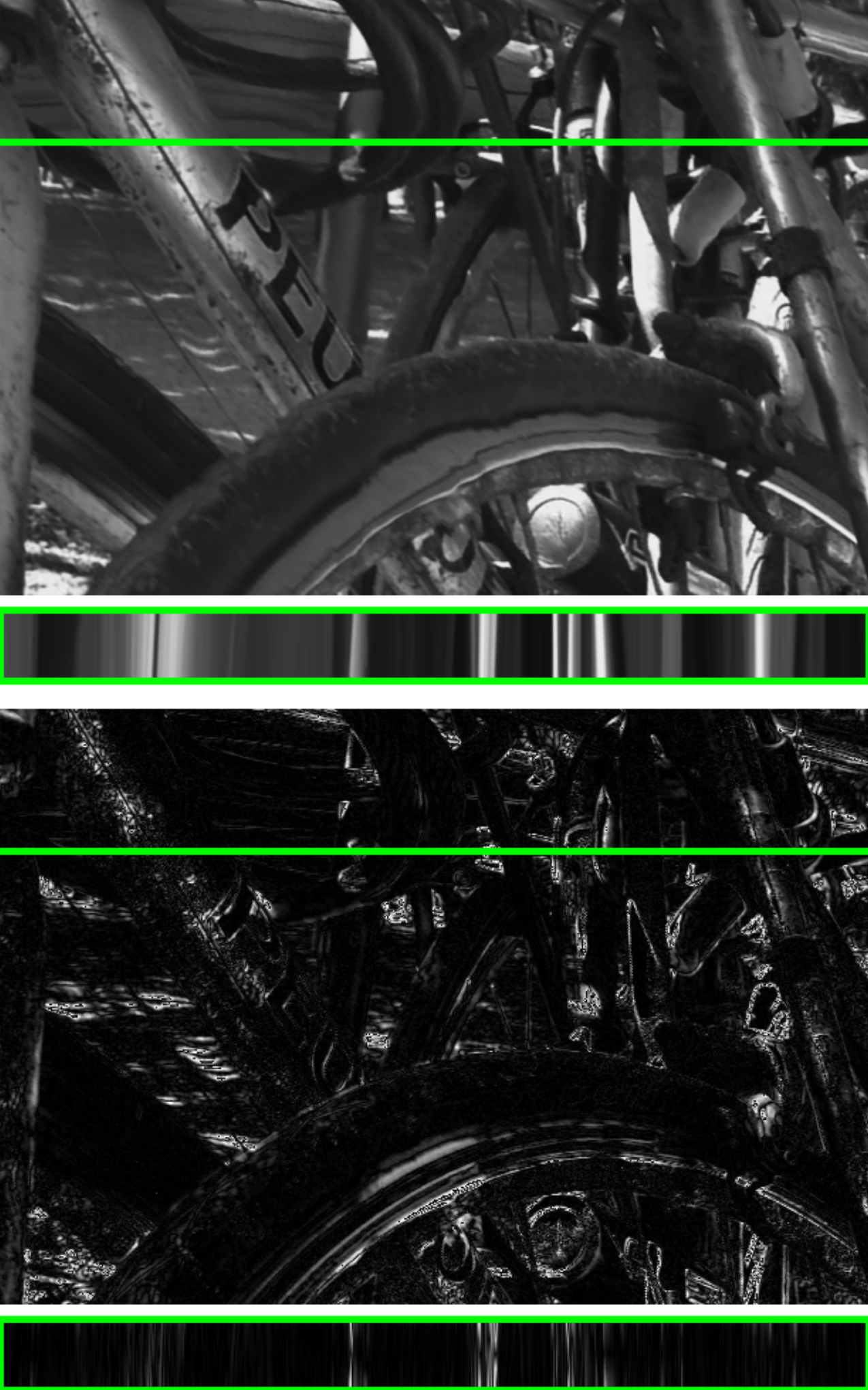} \\ \\
         {\small Ground truth} & {\small Inagaki~\cite{Inagaki_2018_ECCV}} & {\small Guo~\cite{Guo2022TPAMI}} &
         \begin{tabular}{c}
            {\small Mizuno} \\ {{\small (re-trained)}~\cite{Mizuno_2022_CVPR}}
         \end{tabular} &
         {\small Ours} & {\small Ours (center-only)}
    \end{tabular}\\
    \centering
    \caption{Visual results of compressive light-field acquisition for \textit{Lego} (top) and \textit{Bike} (bottom) scenes. For each method, we present top-left view, epipolar plane image (EPI) corresponding to green line, and difference from ground truth ($\times 3$ for better visualization).}
    \label{fig:lego_quant}
}
}
\end{figure*}

\section{Experiments}
\label{sec:results}

We used the BasicLFSR dataset~\cite{BasicLFSR}, which contains 167 light fields collected from five other datasets (EPFL, HCI new, HCI old, INRIA, and Stanford gantry); 144 light fields were assigned for training, and 23 light fields were reserved for testing. As the training samples, we extracted light field patches, each of which had $120\times 120$ pixels and $5 \times 5$ views, from the 144 training light fields. We jointly trained FeatNet and NeRFNet over 120 epochs using Adam optimizer, which took 11.7 hours on an NVIDIA GeForce RTX 3090. To validate the effectiveness of the camera-side coding, we also trained two variants without the coding: ours (no-coding), in which Eq.~(\ref{eq:ordinary}) was used as the camera model, and ours (center-only), in which the central view of the light field was used as the input to FeatNet. 

We evaluated our method from two perspectives: compressive light-field acquisition and continuous light field reconstruction. Please refer to the supplementary video because the visual quality and 3-D consistency of a light field cannot be well presented on paper.

\subsection{Compressive Light-Field Acquisition}

We compared our method with several state-of-the-art methods for compressive light-field acquisition~\cite{Inagaki_2018_ECCV,Guo2022TPAMI,Mizuno_2022_CVPR}. Inagaki et al.'s method~\cite{Inagaki_2018_ECCV} and Guo et al.'s method~\cite{Guo2022TPAMI} are based on coded aperture imaging (Eq.~({\ref{eq:ca}})). These methods were retrained on the BasicLFSR dataset until convergence in the same configuration as ours: $5 \times 5$ views were reconstructed from a single coded image. Mizuno et al.'s method~\cite{Mizuno_2022_CVPR} adopted joint aperture-exposure coding (Eq.~(\ref{eq:ca+ce})). Since Mizuno et al.'s method was originally designed for a moving scene, it produced $5 \times 5$ views over 4 temporal sub-frames. To acquire a light field for a static scene, we took the average over the temporal domain to reduce them into $5 \times 5$ views. We used two models for this method; one was the pre-trained model provided by Mizuno et al.~\cite{Mizuno_2022_CVPR}, and the other was re-trained on the BasicLFSR dataset by ourselves. Note that our method can reconstruct \textbf{continuous} viewpoints, whereas the others~\cite{Inagaki_2018_ECCV,Guo2022TPAMI,Mizuno_2022_CVPR} can reconstruct only \textbf{discretized} $5\times 5$ views. In other words, these methods~\cite{Inagaki_2018_ECCV,Guo2022TPAMI,Mizuno_2022_CVPR} devoted all resources to reconstructing individual $5\times 5$ views rather than a continuous 3-D representation of the target scene.

Table \ref{tab:main_table} summarizes the quantitative scores (PSNR and SSIM; greater is better) for the 23 light fields reserved for testing. These 23 light fields were divided into 5 groups corresponding to the source datasets, and we report the average scores for each group and all the groups. Note that the scores were evaluated only for discrete $5 \times 5$ viewpoints, whereas our method can reconstruct continuous viewpoints. As seen from the table, the camera-side coding had a significant impact on the reconstruction quality. Joint aperture-exposure coding (Mizuno et al.'s~\cite{Mizuno_2022_CVPR}) significantly outperformed aperture coding (Inagaki et al.'s~\cite{Inagaki_2018_ECCV} and Guo et al.'s~\cite{Guo2022TPAMI}). Using joint aperture-exposure coding for image acquisition, our method achieved a reconstruction quality comparable to Mizuno et al.'s~\cite{Mizuno_2022_CVPR} for the discretized $5\times 5$ viewpoints. The quantitative scores for Mizuno et al.'s~\cite{Mizuno_2022_CVPR} were slightly better than those for ours; this is understandable, because Mizuno et al.'s~\cite{Mizuno_2022_CVPR} devoted all the resources to the discrete $5\times 5$ views without considering other viewpoints. Meanwhile, ours (no-coding) and ours (center-only) produced poor results, showing the difficulty of light field reconstruction from a single image without camera-side coding.


\newcommand{\povSize}{0.22\hsize}

\begin{figure*}[t]
{\tabcolsep = 0.8mm
{\renewcommand\arraystretch{0.5}
    \centering
    \begin{tabular}{ccccc}        
          \includegraphics[width = .2\linewidth]{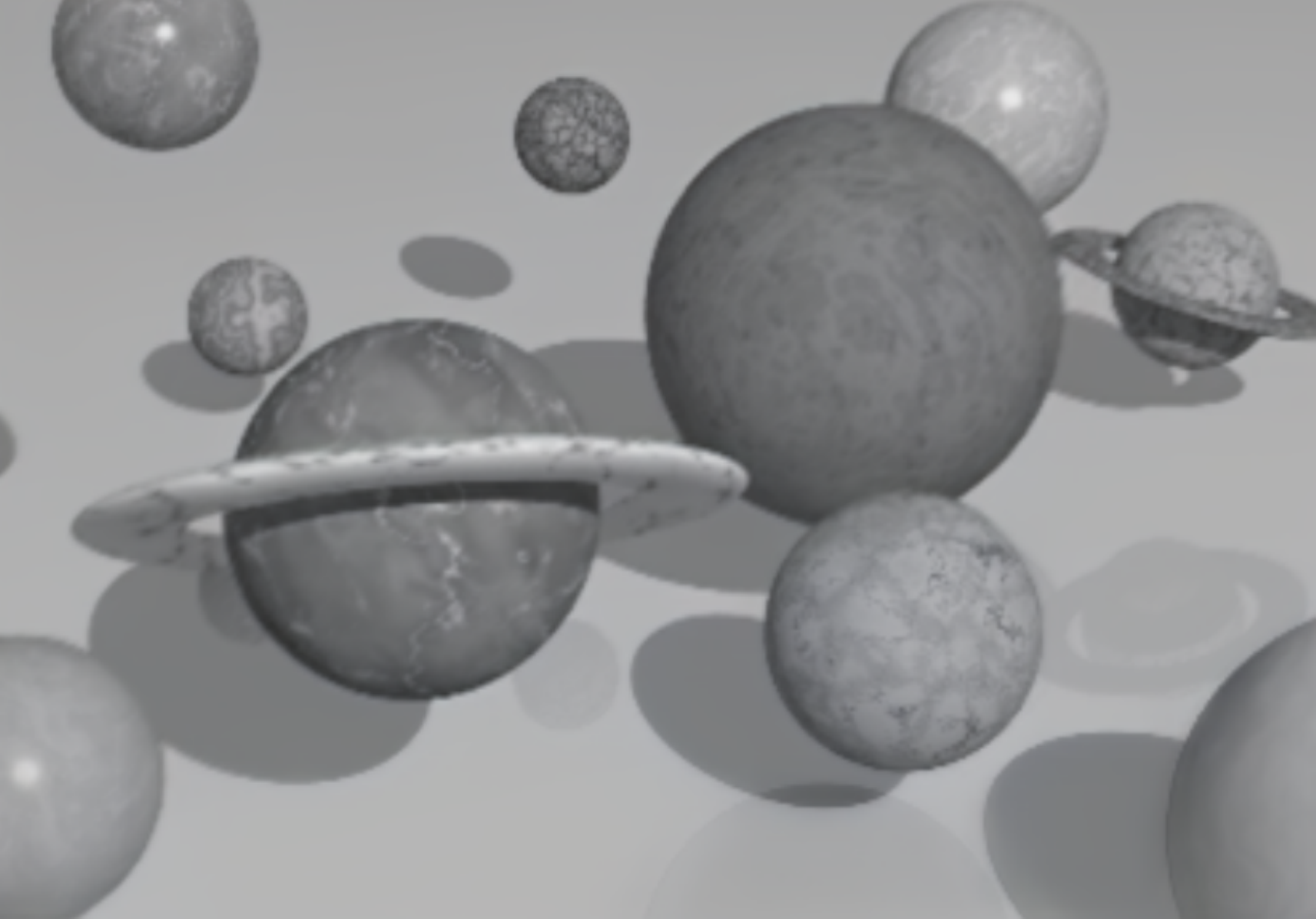} &
          \includegraphics[width = .2\linewidth]{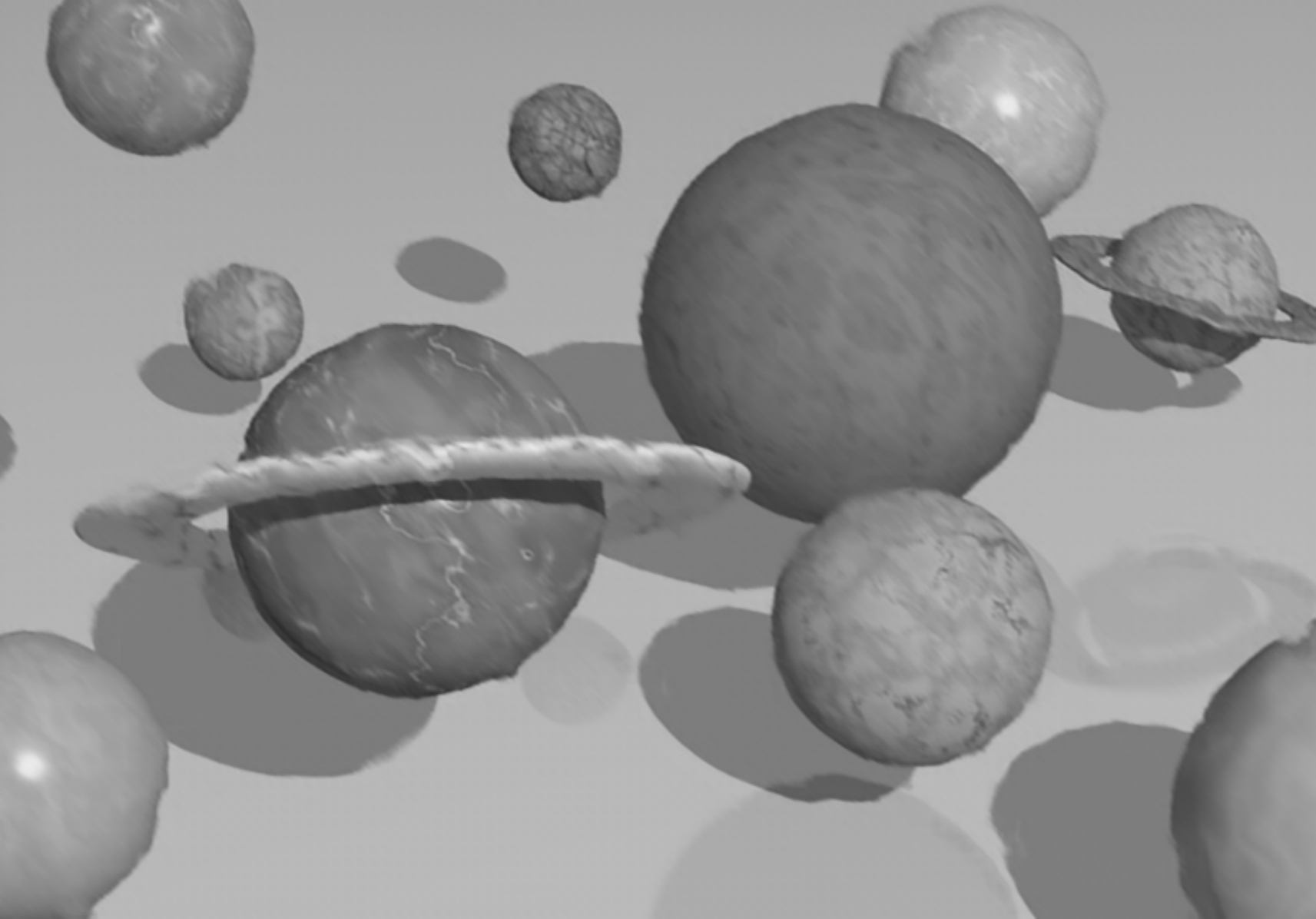} &
          \includegraphics[width = .2\linewidth]{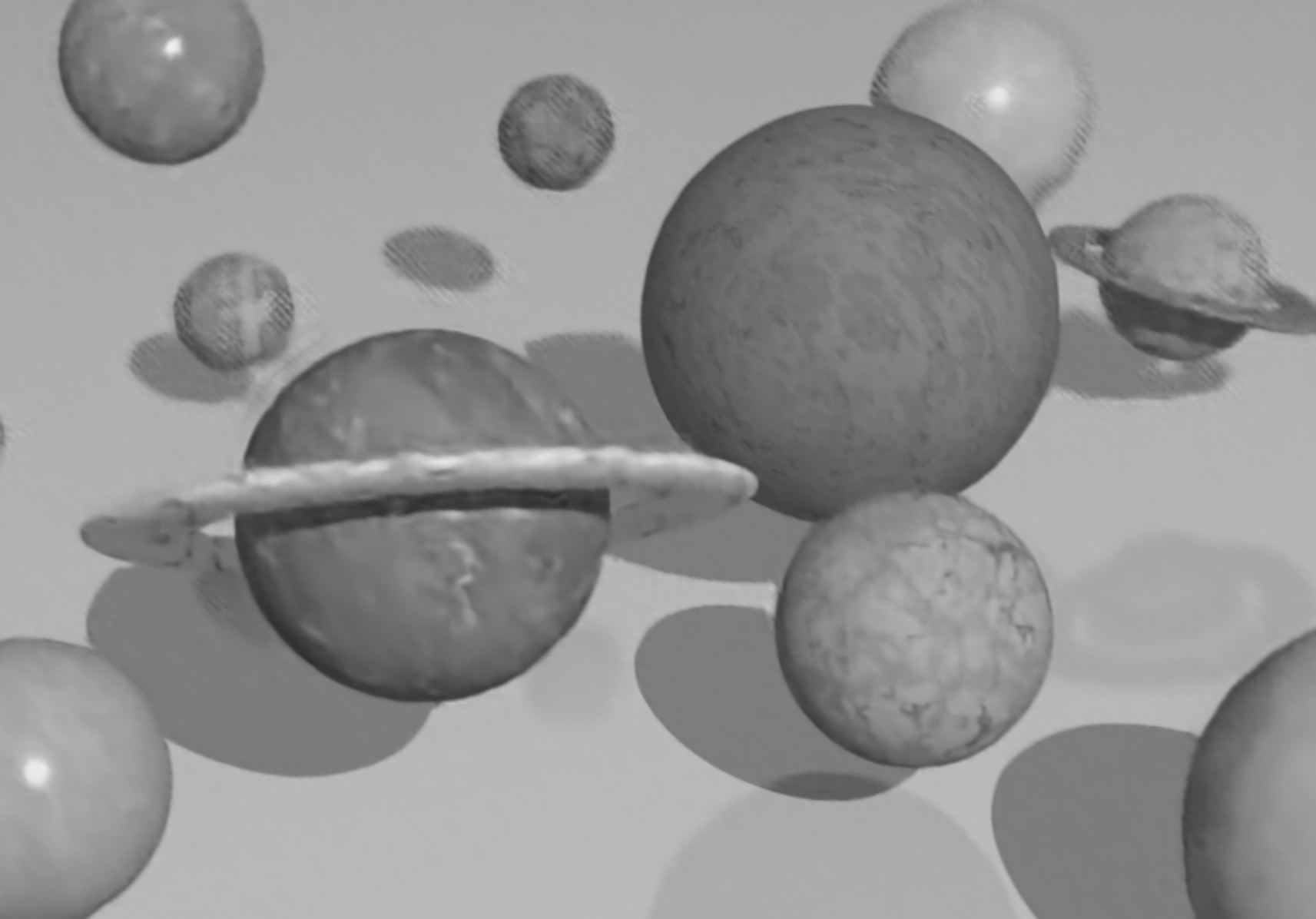} &
          \includegraphics[width = .2\linewidth]{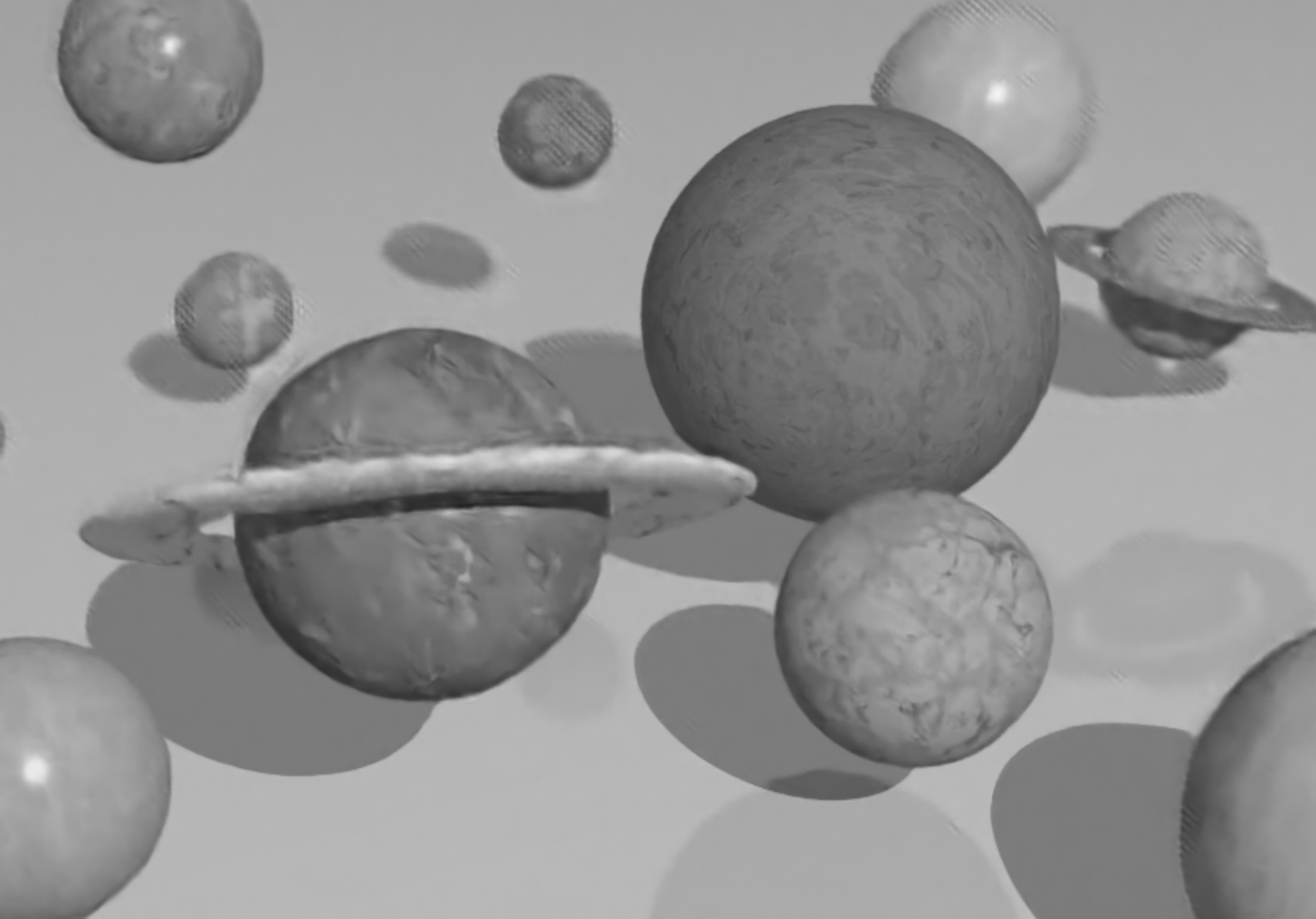} &
            \multirow{2}{*}{
            \begin{minipage}{0.06\hsize}
              \centering
                \vspace{-70pt}
              \includegraphics[width = .9\linewidth]{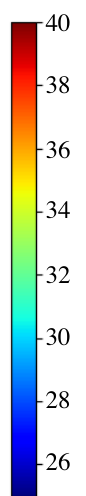}
            \end{minipage}} \\ \\
          \includegraphics[width = .15\linewidth]{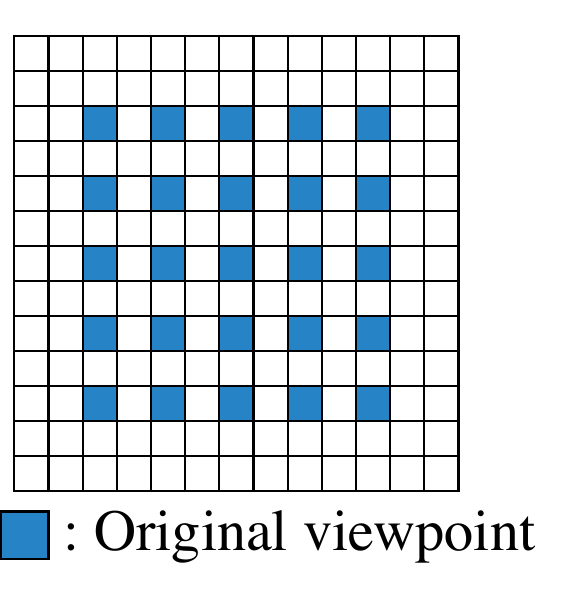} &
          \includegraphics[width = .15\linewidth]{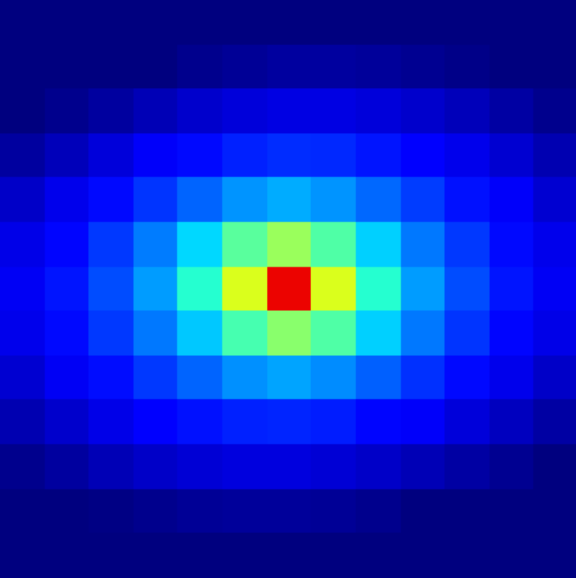} &
          \includegraphics[width = .15\linewidth]{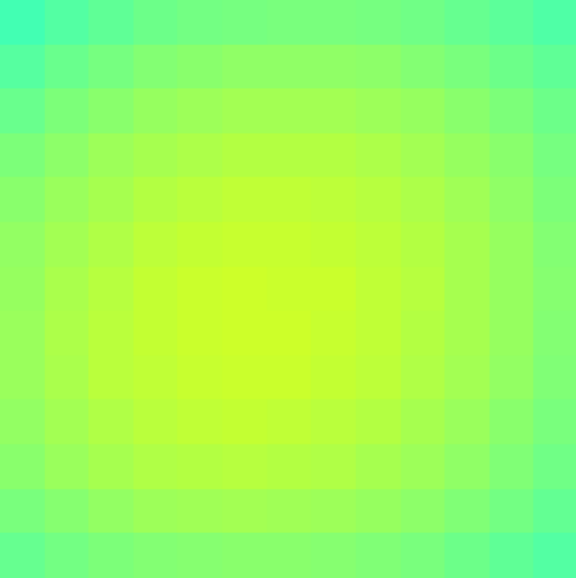} &
          \includegraphics[width = .15\linewidth]{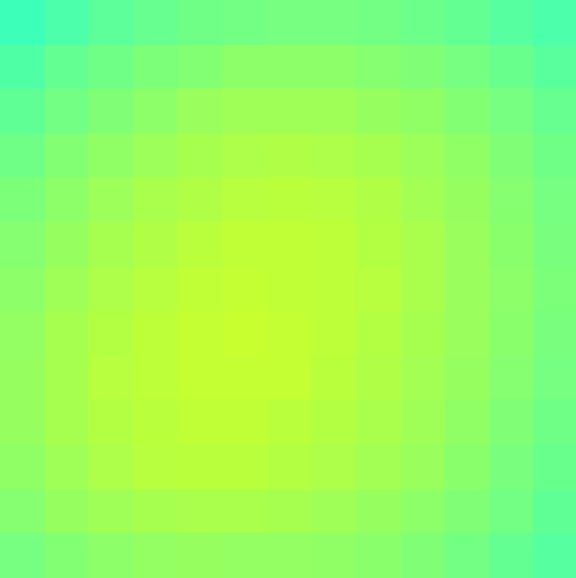} & \\ \\
        MINE & Ours (center-only) & Ours & Mizuno+GeoNeRF & \\ \\
    \end{tabular}\\
    \caption{Continuous light field reconstruction for \textit{Planets} scene. PSNR values for $13 \times 13$ views are presented as heat maps. PSNR values for MINE cannot be computed, and thus, we present viewpoint configuration in column of MINE.}
    \label{fig:povray_result}

    \vspace{5mm}
    \begin{tabular}{cccc}
          \includegraphics[width = .2\linewidth]{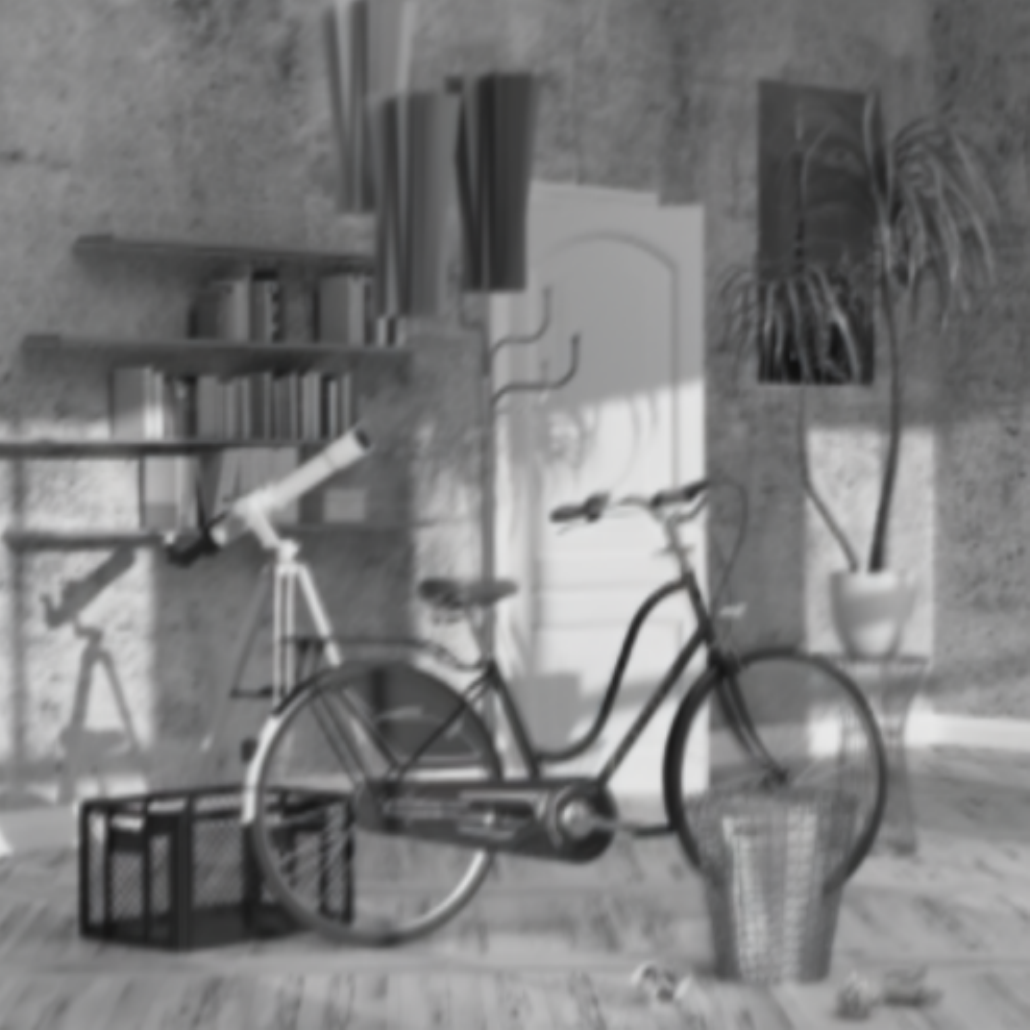} &
          \includegraphics[width = .2\linewidth]{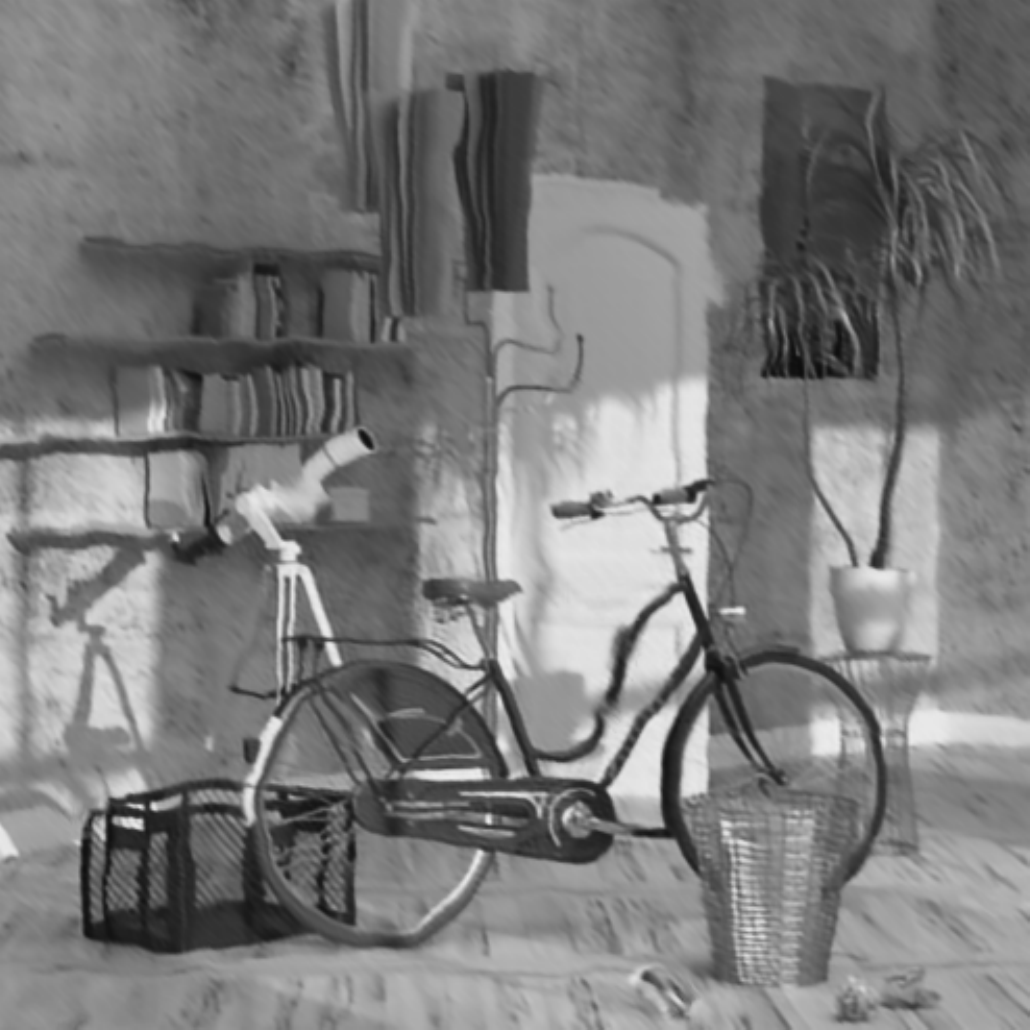} &
          \includegraphics[width = .2\linewidth]{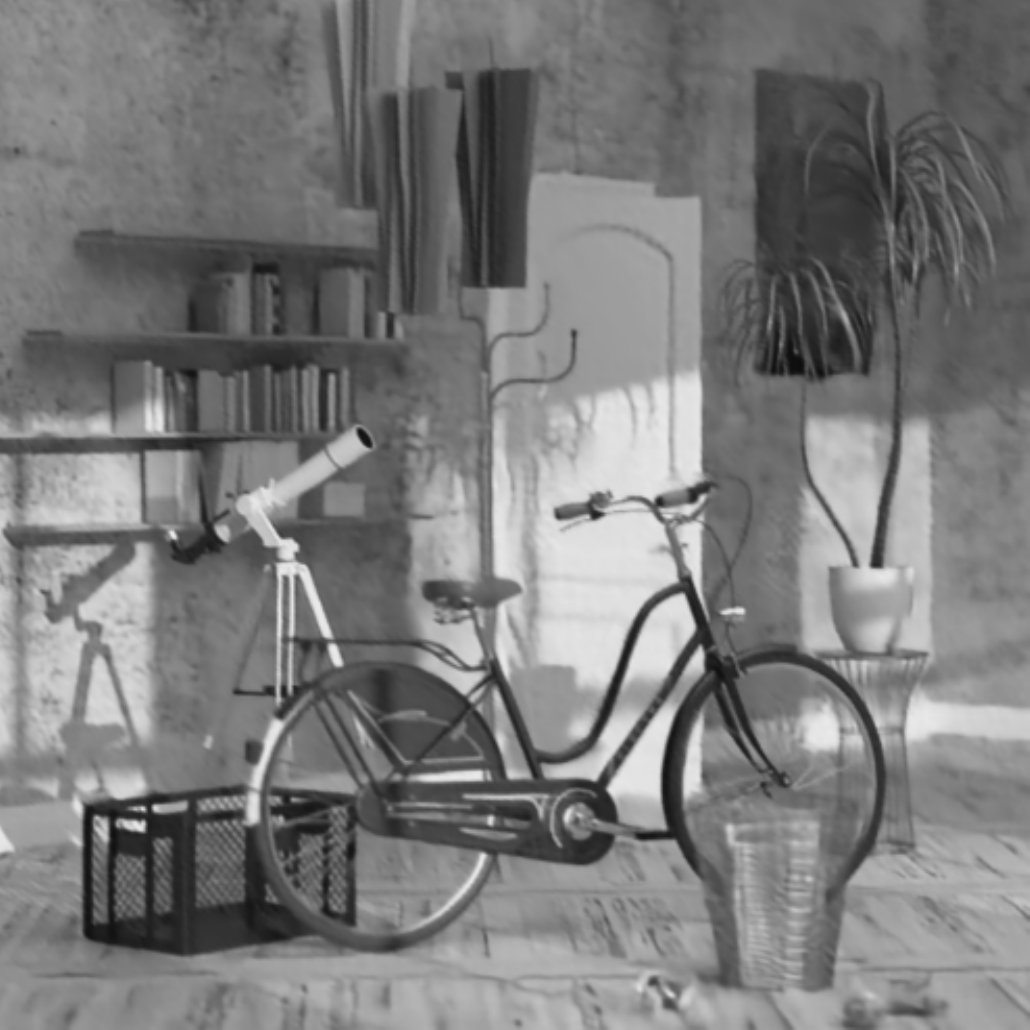} &
          \includegraphics[width = .2\linewidth]{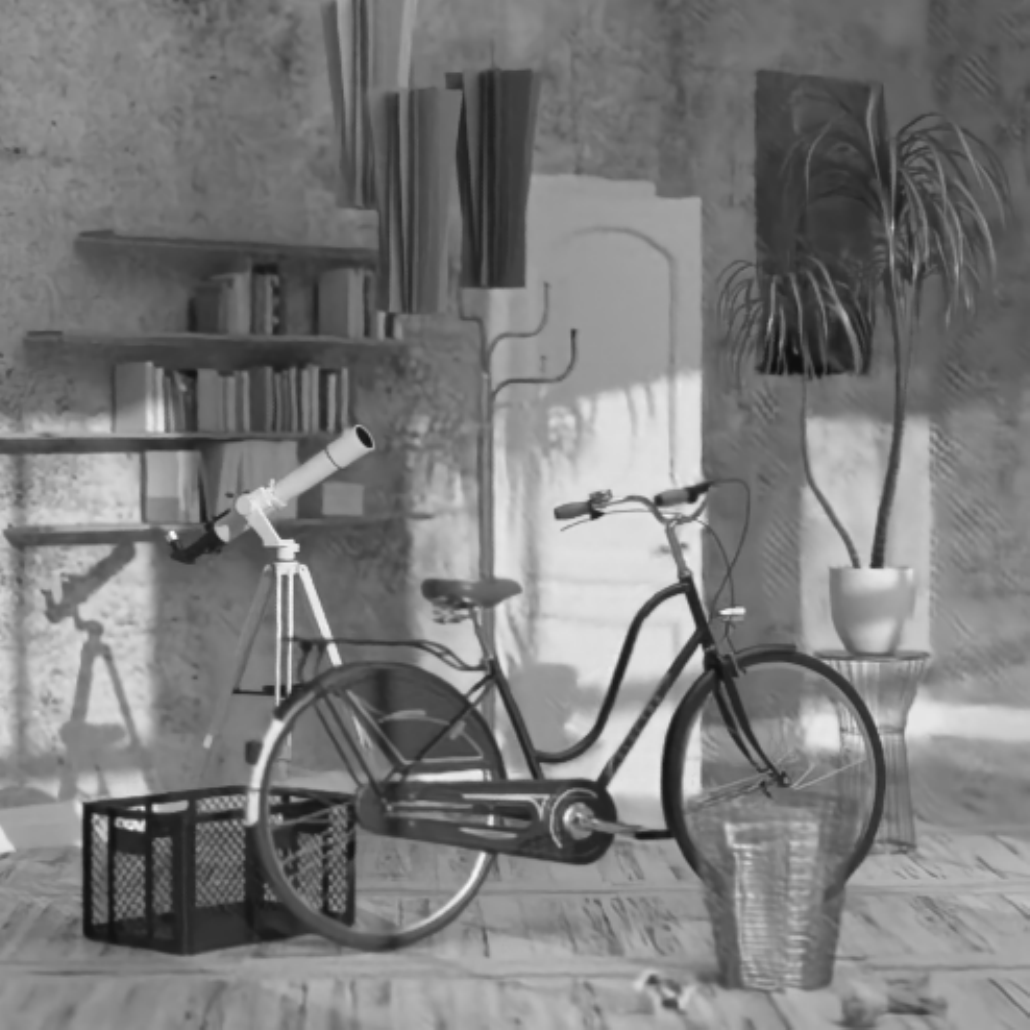} \\
        \includegraphics[width = .2\linewidth]{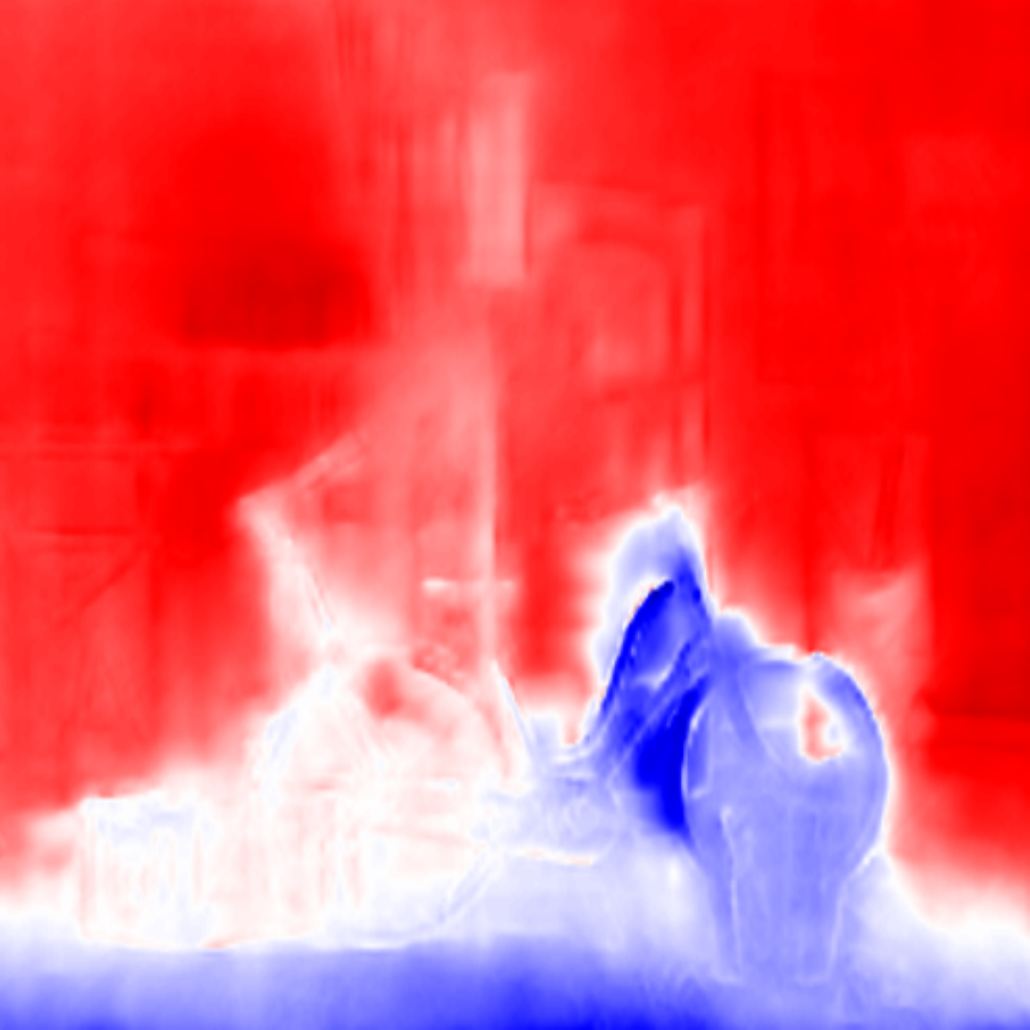} &
          \includegraphics[width = .2\linewidth]{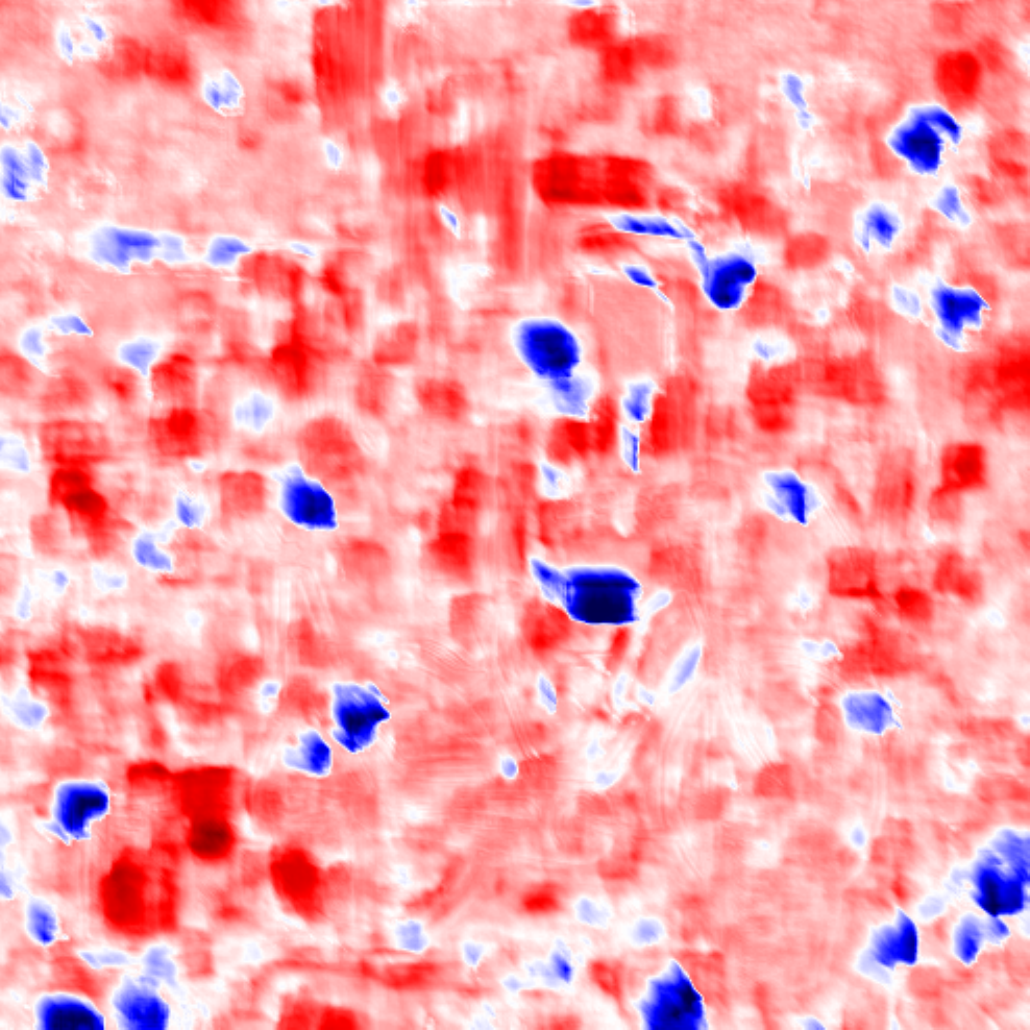} &
          \includegraphics[width = .2\linewidth]{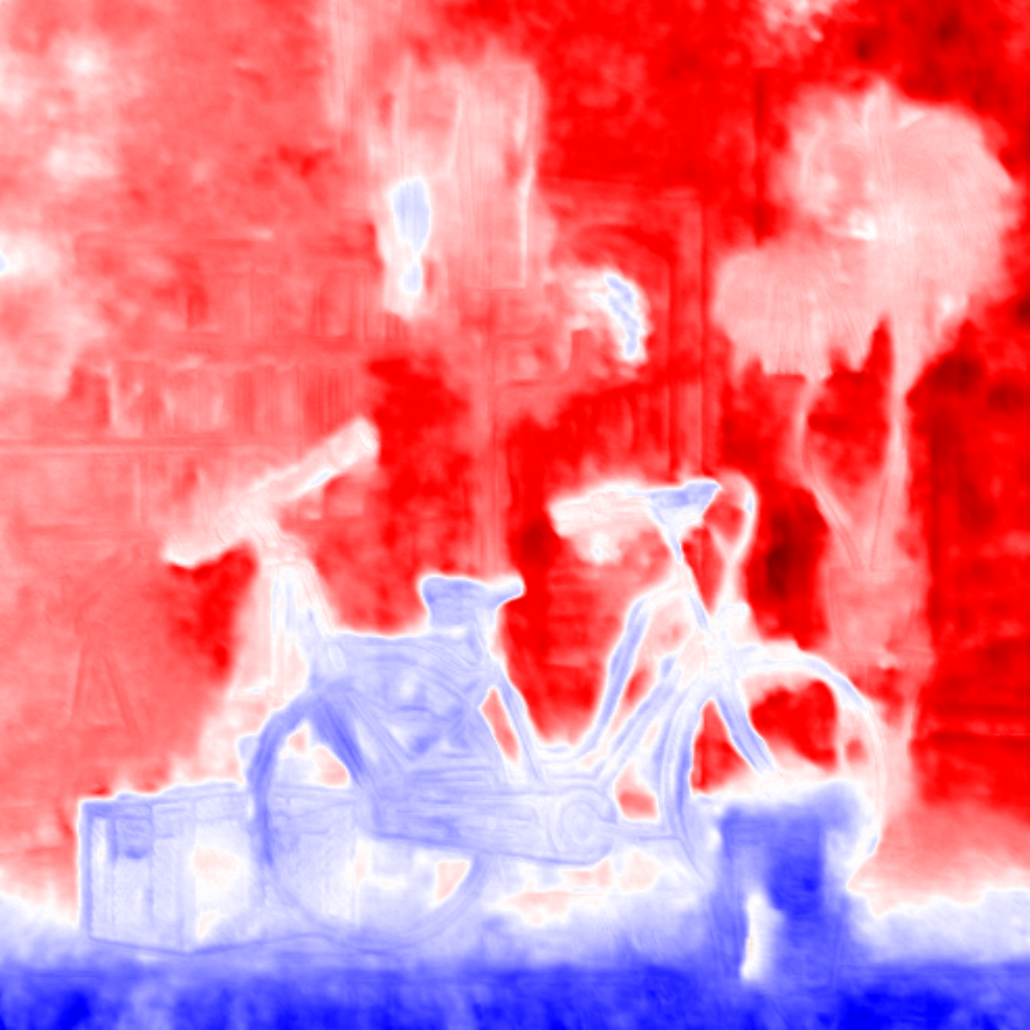} &
          \includegraphics[width = .2\linewidth]{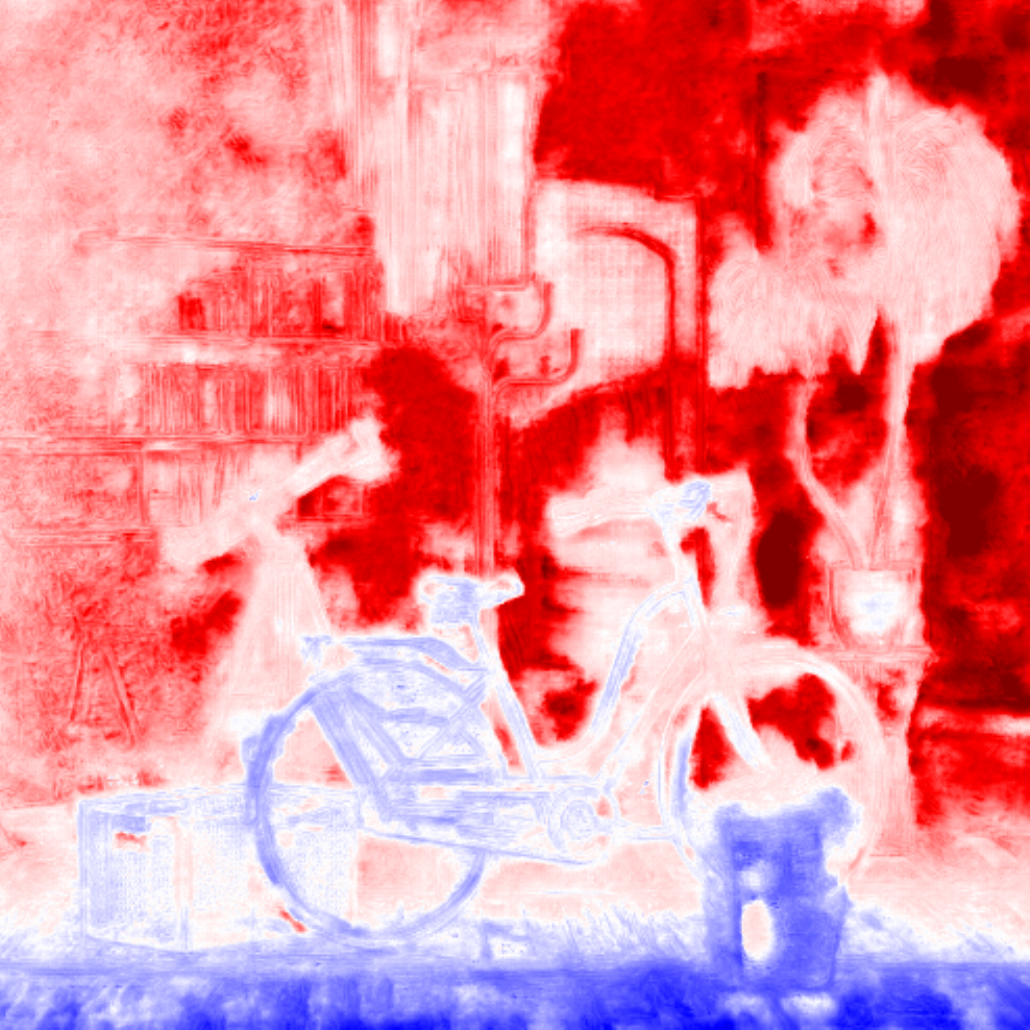} \\
          MINE & Ours (center-only) & Ours & Mizuno+GeoNeRF  \\
    \end{tabular}
    \vspace{5mm}
    \caption{Rendered views (top) and pseudo depth maps (bottom) for \textit{Bicycle} scene. Please refer to supplementary video for better visualization with viewpoint movement and other results.} 
    \label{fig:knight_result}
}
}
\end{figure*}

\subsection{Continuous Light-Field Reconstruction}

We evaluated the capability of our method to reconstruct a continuous light field. For a quantitative evaluation, we used a computer generated scene, \textit{Planets}, provided by Sakai et al.~\cite{Sakai_2020_ECCV}. We used $5 \times 5$ views to compute a coded image by using Eq.~(\ref{eq:ca+ce}), from which we reconstructed a light field with $13 \times 13$ views. As shown in Fig.~\ref{fig:povray_result}, the $13 \times 13$ views include both the original $5 \times 5$ views and interpolated/extrapolated views.  

For comparison, we tested two other methods for continuous light-field reconstruction. The first is \textbf{MINE}~\cite{mine2021ICCV}, a state-of-the-art single-view view synthesis method without test time training. It was reported in \cite{mine2021ICCV} that MINE was superior to several other methods~\cite{Srinivasan2017,tulsiani2018layer,single_view_mpi,Wiles_2020_CVPR}. As the input to MINE, we used the central view of a light field without camera-side coding. The other is \textbf{Mizuno+GeoNeRF}, which is a brute-force concatenation of two state-of-the-art methods; first, by using Mizuno et al.'s method~\cite{Mizuno_2022_CVPR} (the re-trained model), a light field with $5 \times 5$ views was reconstructed from a single coded image; then, these reconstructed views along with the corresponding camera parameters were fed to GeoNeRF~\cite{Geonerf2022CVPR}, a state-of-the-art NeRF-based rendering method without test time training. For both MINE and GeoNeRF, we used pre-trained network models~\footnote{For MINE, we used a model trained on \textit{RealEstate10K} because the ones trained on \textit{Flowers} and \textit{Kitti} were overfitted to a specific category of scenes and performed very poorly for other scenes.} provided by the corresponding authors. Similar to the case with our method, grayscale images were used as input to these methods.

\begin{table*}[t]
  \begin{center}
    \end{center}
    \caption{Comparison of computation times (in seconds) between our method and Mizuno+GeoNeRF measured on NVIDIA GeForce RTX 3090. Although average PSNR scores are almost same, our method is significantly more efficient than Mizuro+GeoNeRF.}
    \label{tab:computation-time}
    \vspace{1mm}
    \begin{tabular}{l||c|ccc}
        & \begin{tabular}{c} Average PSNR [dB] \\ ($13 \times 13$ views) \end{tabular}& \begin{tabular}{c}$5\times5$ view \\ reconstruction \end{tabular} & \begin{tabular}{c} Feature/geometry \\ computation \end{tabular} & \begin{tabular}{c} Rendering \\ (per view) \end{tabular} \\ \hline \hline
        Ours & \textbf{33.01} & - & \textbf{0.14} & \textbf{0.40} \\
        Mizuno+GeoNeRF & 32.93 & 0.12 & 1.58 & 79 \\ \hline

    \end{tabular}
\end{table*}

It should be noted with MINE that single-view view synthesis is scale ambiguous; since only a single view is given as the input, the method cannot know the absolute scale of the scene in principle. Accordingly, we cannot exactly align the rendered viewpoints with those of the target light field.\footnote{We provided MINE with the ground truth camera parameters of the \textit{Planets} scene, but due to the scale difference, the rendered viewpoints were significantly different from those of the target light field.} For this reason, we did not evaluate the quantitative scores for MINE. For visual comparison, we manually configured the viewpoints for MINE to obtain a similarly-looking viewpoint arrangement to the target light field.

The quantitative scores (PSNRs) for the $13 \times 13$ views are visualized as heat maps (each grid corresponds to each viewpoint) in Fig.~\ref{fig:povray_result}. With ours (center-only), the reconstruction quality degraded rapidly as the viewpoint diverged from the central viewpoint, resulting in very poor quality for the marginal viewpoints. This indicates that continuous light-field reconstruction is difficult to achieve without the camera-side coding. In contrast, our method could maintain the quality of all the viewpoints; the reconstruction quality degraded only gradually as the viewpoint diverged from the central viewpoint. Moreover, the reconstruction quality was consistent (not changing abruptly) among the original and the other viewpoints, supporting our claim that our method can reconstruct a light field at continuous viewpoints. The reconstruction quality of our method was almost the same as that of Mizuno+GeoNeRF, which can be regarded as the best achievable quality with the current state-of-the-art. 

Besides the results for the \textit{Planets} scene in Fig.~\ref{fig:povray_result}, we present visual results for the \textit{Knight} scene in Fig.~\ref{fig:knight_result}. Along with the rendered views, we present pseudo depth maps, which can be obtained during the process of volume rendering. Although the depth values were not that accurate in particular for textureless regions, they indicated how each method understood the 3-D structure of the target scene. The reconstruction quality of MINE was obviously poor. The pseudo depth map indicates that MINE seriously failed to reconstruct the 3-D shape of the target scene. As the viewpoint moved, we observed that the object shapes were distorted significantly. These observations indicate that light field reconstruction from a normal (uncoded) image alone is an ill-posed problem. For the same reason, ours (center-only) also resulted in significantly poor visual quality. Meanwhile, thanks to the camera-side coding, our method achieved visually-convincing and 3-D-consistent reconstruction. The reconstruction quality of our method was comparable to that of Mizuno+GeoNeRF. 

We finally mention the computational advantage of our method over Mizuno+GeoNeRF. We measured the computational times for these methods using the \textit{Planets} scene on an NVIDIA GeForce RTX 3090. As shown in Table \ref{tab:computation-time}, our method was substantially more efficient than Mizuno+GeoNeRF. Our method derived a feature volume directly from the coded image with a small amount of computational time (0.14~sec), which was comparable to that for the discretized light field reconstruction in Mizuno+GeoNeRF (0.12~sec). Using the feature volume, our method can perform rendering from continuous viewpoints. Meanwhile, Mizuno+GeoNeRF required another 1.58~sec to compute the geometry before the rendering process.  Moreover, the neural rendering process of our method was much more light-weight than that for GeoNeRF; 0.40~sec/view with our method against 79~sec/view with Mizuno+GeoNeRF. Despite being very efficient, our method achieved a rendering quality comparable to (slightly better than, for this scene) Mizuno+GeoNeRF. This can be attributed to our unified design of the feature extraction and neural rendering processes, which enabled end-to-end optimization over the entire pipeline. 

\section{Conclusion}

We proposed a method for reconstructing a continuous light field of a target scene from a single observed image. To this end, we integrated two state-of-the-art techniques into a unified and end-to-end trainable pipeline: joint aperture-exposure coding for compressive light-field acquisition, and a NeRF-based representation for high-quality rendering from continuous viewpoints. Experimental results showed that our method can reconstruct continuous light fields accurately and efficiently without any test time optimization. To our knowledge, this is the first work to bridge two worlds: camera design for efficiently acquiring 3-D information and neural rendering for high-quality view synthesis from continuous viewpoints. 

\textbf{Limitations and future work}. Although joint aperture-exposure coding is quite effective at obtaining the 3-D information of a target scene, it requires an elaborate hardware implementation. Considering the existence of imaging hardware, we assumed that the target light fields were grayscale. Extension to RGB colors remains as future work, which should involve further hardware development. Moreover, we kept our networks simple for computational efficiency, which put an upper-bound on the reconstruction quality. Using deeper networks would improve the reconstruction quality at the cost of increased computational cost. Finding the balance between quality and efficiency is also an important future direction.

\bibliographystyle{reference/IEEEbib}
\bibliography{reference/refs}


\begin{IEEEbiography}[{\includegraphics[draft=false,width=1in,height=1.25in,clip,keepaspectratio]{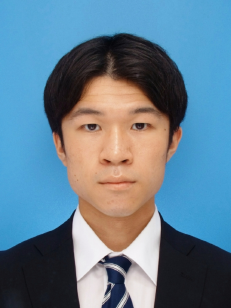}}]
{Yuya Ishikawa} received his B.E. in electrical engineering from Nagoya University, Japan, in 2021. He is currently a graduate student at the Graduate School of Engineering, Nagoya University, Japan. His research topics are light-field and neural representation.
\end{IEEEbiography}

\begin{IEEEbiography}[{\includegraphics[draft=false,width=1in,height=1.25in,clip,keepaspectratio]{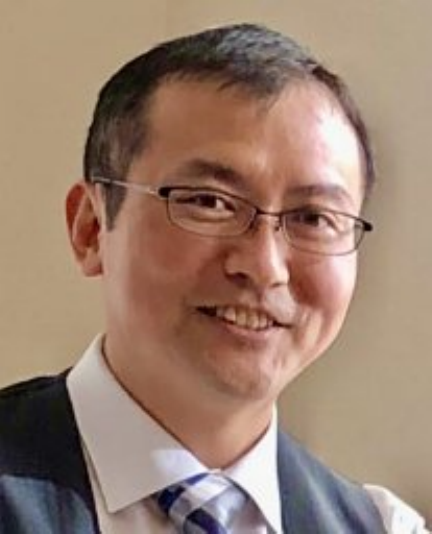}}]
{Keita Takahashi} received B.E., M.S., and Ph.D. degrees in information and communication engineering from the University of Tokyo, Japan, in 2001, 2003, and 2006. He was a project assistant professor at the University of Tokyo from 2006 to 2011 and an assistant professor at the University of Electro-Communications from 2011 to 2013. Since 2013, he has been with the Graduate School of Engineering, Nagoya University, as an associate professor. His research interests include image processing, computational photography, and 3D displays. He is a member of the IEEE Computer Society.
\end{IEEEbiography}

\begin{IEEEbiography}[{\includegraphics[draft=false,width=1in,height=1.25in,clip,keepaspectratio]{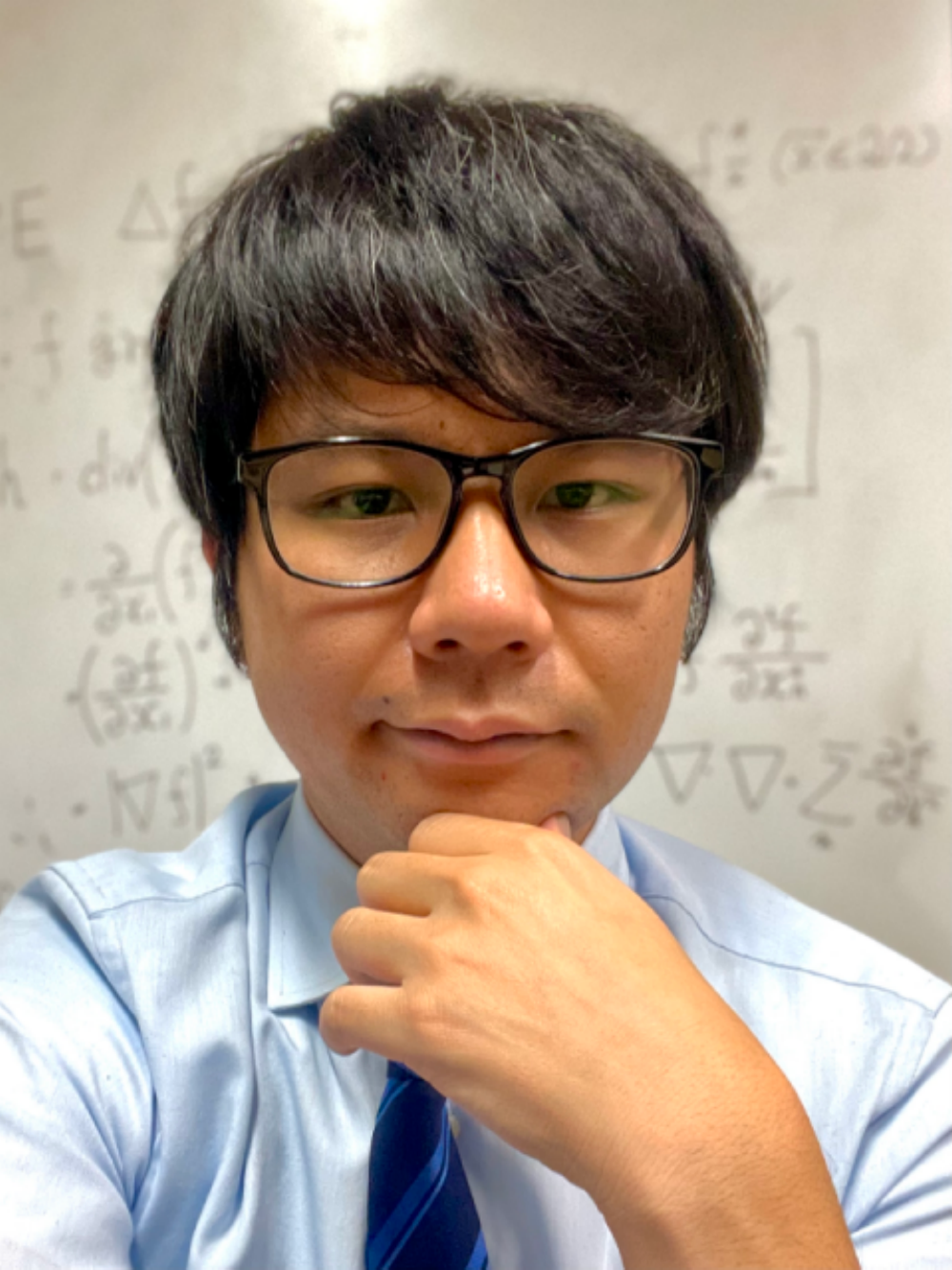}}]
{Chihiro Tsutake} received B.E., M.E., and Ph.D. degrees from the University of Fukui, Japan, in 2015, 2017, and 2020. Since 2020, he has been with the Graduate School of Engineering, Nagoya University, as an assistant professor. His research interests include 3D image processing, image coding, optical systems, and applied mathematics.
\end{IEEEbiography}

\begin{IEEEbiography}[{\includegraphics[draft=false,width=1in,height=1.25in,clip,keepaspectratio]{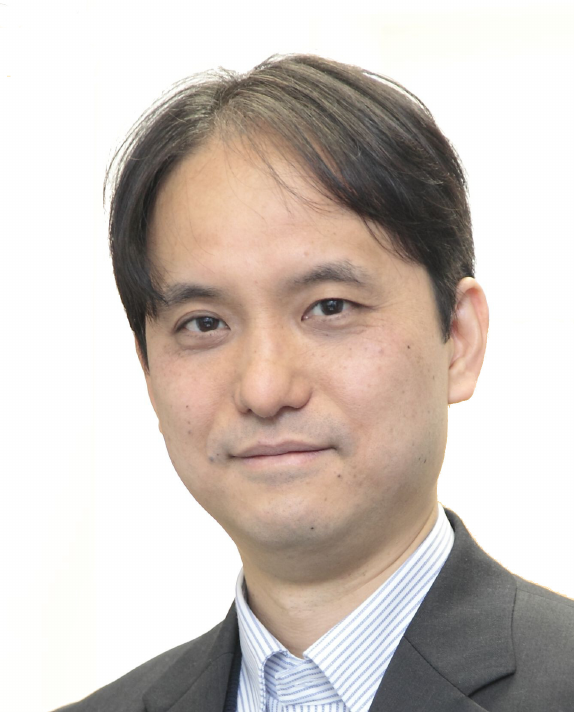}}]
{Toshiaki Fujii} received B.E., M.E., and Dr.E. degrees in electrical engineering from the University of Tokyo, Japan, in 1990, 1992, and 1995. In 1995, he joined the Graduate School of Engineering, Nagoya University, where he is currently a professor. From 2008 to 2010, he was with the Graduate School of Science and Engineering, Tokyo Institute of Technology. From 2019 to 2021, he also served as a senior science and technology policy fellow of the Cabinet Office, Government of Japan. His current research interests include multidimensional signal processing, multi-camera systems, multi-view video coding and transmission, free-viewpoint video, and their applications. He is a member of the IEEE Signal Processing Society, the ISO/IEC JTC1/SC29/WG4, WG1 (MPEG-I Visual, JPEG) standardization committee of Japan, the Information Processing Society of Japan, and the Institute of Image Information and Television Engineers of Japan.

\end{IEEEbiography}
\EOD

\end{document}